\documentclass[sageh,Afour,times]{sagej}  % Comment this line out if you need a4paper

\usepackage[colorlinks,bookmarksopen,bookmarksnumbered,citecolor=red,urlcolor=red]{hyperref}

\usepackage{bm}
\usepackage{graphics} % for pdf, bitmapped graphics files
\usepackage{amsmath} % assumes amsmath package installed
\usepackage{amssymb}  % assumes amsmath package installed
\usepackage{multicol}
\usepackage{lipsum}
\usepackage{xcolor}
\usepackage{graphicx}
\usepackage{subcaption}

\allowdisplaybreaks

\newcommand{\R}{\mathbb{R}}

\newcommand{\pyrobocop}{\textsc{PyRoboCOP}}
\newcommand{\adolc}{\textsf{ADOL-C}}
\newcommand{\ipopt}{\textsf{IPOPT}}
\newcommand{\casadi}{\textsf{CasADi}}
\newcommand{\pyomo}{\textsf{Pyomo}}

\newcommand{\mysub}[2]{[#1]_{#2}}
\newcommand{\setcompl}{{\cal L}}

\newcommand{\lb}[1]{\underline{#1}}
\newcommand{\ub}[1]{\overline{#1}}
\newcommand{\nfe}{N_e}
\newcommand{\ncoll}{N_{c}}
\newcommand{\setnfe}{{\cal N}_e}
\newcommand{\setncoll}{{\cal N}_c}

\newcommand{\setmodes}{{\cal M}}

\newtheorem*{remark}{Remark}

\begin{document}
\runninghead{\pyrobocop}

\title{\LARGE \bf
\pyrobocop : \underline{Py}thon-based \underline{Robo}tic \underline{C}ontrol \& \underline{O}ptimization \underline{P}ackage 
for Manipulation and Collision Avoidance
}

\author{Arvind U. Raghunathan$^{1}$, Devesh K. Jha$^{1}$and Diego Romeres$^{1}$% <-this % stops a space
% <-this % stops a space
% \thanks{$^1$All authors are with Mitsubishi Electric Research Laboratories (MERL), Cambridge, MA 02139. Email--{ \tt\small \{raghunathan,jha, romeres\}@merl.com}}
}

\affiliation{
\affilnum{1}All authors are with Mitsubishi Electric Research Laboratories (MERL), Cambridge, MA 02139. Email--{ \tt\small \{raghunathan,jha, romeres\}@merl.com}}

\begin{abstract}
\pyrobocop\, is a lightweight Python-based package for control and optimization of robotic systems described by nonlinear Differential Algebraic Equations (DAEs).  In particular, the package can handle systems with contacts that are described by complementarity constraints and provides a general framework for specifying obstacle avoidance constraints. The package performs direct transcription of the DAEs into a set of nonlinear equations by performing orthogonal collocation on finite elements.  The resulting optimization problem belongs to the class of Mathematical Programs with Complementarity Constraints (MPCCs).  MPCCs fail to satisfy commonly assumed constraint qualifications and require special handling of the complementarity constraints in order for NonLinear Program (NLP) solvers to solve them effectively.  \pyrobocop\, provides automatic reformulation of the complementarity constraints that enables NLP solvers to perform optimization of robotic systems. The package is interfaced with \adolc\,~\citep{ADOLC} for obtaining sparse derivatives by automatic differentiation and \ipopt\,~\citep{IPOPT} for performing optimization.  We demonstrate the effectiveness of our approach in terms of speed and flexibility. We provide several numerical examples for several robotic systems with collision avoidance as well as contact constraints represented using complementarity constraints.  We provide comparisons with other open source optimization packages like \casadi\ and \pyomo .

\end{abstract}

\maketitle
% \thispagestyle{plain}
% \pagestyle{plain}

%%%%%%%%%%%%%%%%%%%%%%%%%%%%%%%%%%%%%%%%%%%%%%%%%%%%%%%%%%%%%%%%%%%%%%%%%%%%%%%%

\section{Introduction}\label{sec:intro}
Most robotic applications are characterized by presence of constrained environments while dealing with challenging underlying phenomena like unilateral contacts, frictional contacts, impact and deformation~\citep{mason2018toward}. These phenomena are very challenging to understand and represent mathematically. However, these are central to most of the manipulation problems and is important to model these phenomena. This has led to the development of a number physics engines to model contact-rich multi-body interaction~\citep{todorov2012mujoco}. The main motivation of these engines is to allow faster simulation of multi-body dynamics and thus, facilitate real-time model-based control for contact-rich tasks. However, in order to achieve autonomous robotic manipulation, robots must have the tools to reason about environmental contact and discover optimal behavior. This requires development of easy-to-use, flexible optimization packages that can allow solution to mathematical programs that represent such constrained problems arising in the presence of environmental contacts. It is noted that a lot of research in the machine learning community has been fueled by the availability of open-source software that allows quick prototyping of machine learning models~\citep{abadi2016tensorflow}. Our work in this paper is motivated by the understanding that, in the near future, robots could be equipped with tools that allow specification of dynamics (e.g., the physics engines) as well as high-performance optimization routines that allow solution to mathematical programs for model-based optimization and control~\citep{9387127}. With this motivation, we present a python-based robotic control and optimization package (called \pyrobocop) that allows solution to a large class of mathematical programs with nonlinear constraints. The current paper and package only considers systems which can be represented by DAEs. Integration with physics engines is left as a future work as that requires additional development so as to expose the contact model specifications to \pyrobocop\ for optimal performance and will be provided in future.

Contact-rich robotic manipulation tasks are mostly characterized by non-linear geometric constraints as well as constraints imposed by contacts that are modeled as complementarity constraints. Obtaining a feasible, let alone an optimal trajectory, can be challenging for such systems.  An effective integration of the high-level trajectory planning in configuration space with physics-based dynamics is necessary in order to obtain optimal performance of such robotic systems. To the best of author's knowledge, none of the existing python-based open-source optimization packages can provide support for trajectory optimization with support for complementarity constraints that arise from contact-rich manipulation and the easy specification of obstacle avoidance constraints. Such optimization capability is, however, highly desirable to allow easy solution to optimization problems for a large-class of contact-rich robotic systems. 
%This restricts the use of several state-of-the-art (SOTA) algorithms to some case studies and makes benchmarking and comparisons between different SOTA algorithms very challenging. 

In this paper, we present \pyrobocop\, -- a lightweight but powerful Python-based package for control and optimization of robotic systems.  A key contribution of our paper is that we present a novel complementarity-based formulation for modeling collision avoidance. Our formulation is differentiable even when the obstacles or objects are modeled as polytopes. The proposed formulation allows us to handle contact and collision avoidance in an unified manner. \pyrobocop\ uses \adolc\,~\citep{ADOLC} and \ipopt\,~\citep{IPOPT} at its backend for automatic differentiation and optimization respectively.
The main features of the package are:
\begin{itemize}
    \item Direct transcription by orthogonal collocation on finite elements
    \item Contact modeling by complementarity constraints
    \item Obstacle avoidance modeling by complementarity constraints
    \item Automatic differentiation for sparse derivatives
    %\item RRT-based initialization
    \item Support for minimum time problems
    \item Support for optimization over fixed mode sequence problems with unknown sequence time horizons
%    \item Optimization with ML models
\end{itemize}
The features described above should convince the reader that \pyrobocop\, addresses the identified gaps in existing software for optimization of robotic systems. By bringing together \adolc\,~\citep{ADOLC} and \ipopt\,~\citep{IPOPT} we believe that \pyrobocop\ would be very useful for real-time model-based control of robotic systems. Codes and instructions for installing and using \pyrobocop\ will be provided in the camera-ready version of the paper.

\textbf{Contributions.} The main contributions of the paper are:
\begin{enumerate}
    \item We present a python-based package for optimization and control of a large class of robotic systems with contact and collision constraints.
    \item We present a novel formulation for trajectory optimization in the presence of obstacles using complementarity constraints, thus allowing to solve for trajectory optimization of contact-rich systems in the presence of obstacles.
    \item We evaluate our proposed package, \pyrobocop\ , over a range of different dynamical systems and also provide some comparison with SOTA optimization packages for performance and efficiency.
\end{enumerate}

The rest of the paper is structured as follows. In Section~\ref{sec:related_work}, we contrast \pyrobocop\ to existing trajectory optimization techniques and present related works. Section~\ref{sec:probdesc} describes the dynamic optimization problem solved by \pyrobocop\ and specifies the corresponding mathematical program obtained on collocation using finite elements. In Section~\ref{sec:prob_instants}, we describe several variants of discontinuous problems that could be solved by \pyrobocop. Section~\ref{sec:software_description} presents the user interface for \pyrobocop.\ Finally, we show results on a range of different dynamical systems in Section~\ref{sec:results}. Conlusions and future work are summarized in Section~\ref{sec:conclusions}. 

\subsection{Notation} The set of real numbers is denoted by $\R$. Given a vector $u \in \R^n$, $\mysub{x}{i}$ refers to the $i$-th element of the vector.

\section{Related Work}~\label{sec:related_work}
Our work is closely related to various optimization techniques proposed to solve contact-implicit trajectory optimization. Some related examples could be found in~\cite{manchester2020variational, patel2019contact, erez2012trajectory, mordatch2012contact, mordatch2012discovery, YuntGlocker05,YuntGlocker07,Yunt11}. In a more general setting, our work is related to trajectory optimization in the presence of non-differentiable constraints. These problems are common in systems with constraints like non-penetrability~\citep{posa2014direct}, minimum distance (e.g., in collision avoidance)~\citep{zhang2018autonomous}, or in some cases robustness constraints~\citep{kolaric2020local}. 

Some of the existing open-source software for dynamic optimization are Optimica~\citep{Optimica2010},  ACADO Toolkit~\citep{ACADO2011}, TACO~\citep{TACO2013}, pyomo.dae~\citep{PYOMO.DAE2018}, Drake~\citep{drake} and CasADi~\citep{CASADI2019}.   All of the citepd software leverage automatic differentiation to provide the interfaced NLP solvers with first and second-order derivatives.  However, these software do not provide any support for handling contact-based manipulation and obstacle-avoidance which are key requirements in robotic applications. More recently, some packages have been proposed to perform contact-rich tasks in robotics~\citep{9196673}. However, the solver proposed in~\citep{9196673} uses DDP-based~\citep{murray1979constrained} techniques which suffer from sub-optimality and difficulty in constraint satisfaction. Compared to most of the other techniques in open literature, the proposed optimization framework provides the following novelties and advantages.
\begin{enumerate}
    \item Automatic transcription of DAEs to nonlinear equations using collocation on finite elements.
    \item Automatic formulation of collision avoidance between pairs of objects with minimal user input.
    \item Multiple formulations for handling complementarity constraints using NLP solvers robustly. 
\end{enumerate}

The optimization method presented in our work is most closely related to the direct trajectory optimization method for contact-rich systems earlier presented in~\citep{YuntGlocker05,YuntGlocker07,Yunt11}. The authors pose contact dynamics as a measure differential inclusion and employ an augmented Lagrangian to solve the resulting complementarity constrained optimization problem. \cite{posa2014direct} handle the complementarity constraint by relaxing to an inequality and solving using an active-set solver. In an analogous manner, other optimization packages like \casadi\ and \pyomo\ can also be extended for the solution to trajectory optimization in presence of complementarity constraints through a similar reformulation of the complementarity constraints. We provide an adaptive approach for relaxing the complementarity constraints. Further, we also provide a novel formulation for trajectory optimization in the presence of minimum distance constraints for collision avoidance. To the best of our knowledge, there is no other existing open-source, python-based optimization toolbox that can handle constraints arising due to frictional contact interaction and collision avoidance. Consequently, none of the existing trajectory optimization techniques and toolboxes could be used for solving frictional interaction in the presence of additional obstacles.

\section{Problem Description}\label{sec:probdesc}

\pyrobocop\, solves the dynamic optimization problem
\begin{subequations}
\begin{align}
    \min\limits_{x,y,u,p}        &\, \int\limits_{t_0}^{t_f} c(x(t),y(t),u(t),p) dt + \phi(x(t_f),p)\label{dynopt:obj} \\
    \text{s.t.} &\, f(\dot{x}(t),x(t),y(t),u(t),p) = 0,\, x(t_0) = x_0 \label{dynopt:daeqn} \\
                &\, (\mysub{y(t)}{\sigma_{l,1}} - \nu_{l,1}) (\mysub{y(t)}{\sigma_{l,2}} - \nu_{l,2}) = 0 \,\forall\, l \in \setcompl \label{dynopt:ceqn} \\
                &\, \lb{x} \leq x(t) \leq \ub{x}, \lb{y} \leq y(t) \leq \ub{y}, \lb{u} \leq u(t) \leq \ub{u} \label{dynopt:bnds}
\end{align}\label{dynopt}
\end{subequations}
where $x(t) \in \R^{n_x}$, $y(t) \in \R^{n_y}$, $u(t) \in \R^{n_u}$, $\dot{x}(t) \in \R^{n_x}$, $p \in \R^{n_p}$  are the differential, algebraic, control, time derivative of differential variables and time-invariant parameters respectively. The function $\phi : \R^{n_x+n_p} \rightarrow \R$ represents Mayer-type objective function~\citep{BrysonHo} term and is not a function of the entire trajectory.  In addition, $\underline{x}, \overline{x}$, $\underline{y}, \overline{y}$, $\underline{u}, \overline{u}$ are the lower and upper bounds on the differential, algebraic and control variables. The initial condition for the differential variables is $x_0$.    
Constraints~\eqref{dynopt:daeqn}-\eqref{dynopt:ceqn} are the Differential Algebraic Equations (DAEs) modeling the dynamics of the system with $f : \R^{2n_x+n_y+n_u} \rightarrow \R^{n_x+n_y-n_c}$ with $n_c = |\setcompl|$. 
Each $l \in \setcompl$ defines a pair of indices $\sigma_{l,1},\sigma_{l,2} \in \{1,\ldots,n_y\}$ that specifies the complementarity constraint between the algebraic variables $\mysub{y(t)}{\sigma_{l,1}}$ and $\mysub{y(t)}{\sigma_{l,2}}$.  In~\eqref{dynopt:ceqn} $\nu_{l,1}, \nu_{l,2}$ correspond to either the lower or upper bounds on the corresponding algebraic variables.  For example, if they are set respectively to the lower and upper bounds of corresponding algebraic variables then~\eqref{dynopt:ceqn} in combination with the bounds~\eqref{dynopt:bnds} model the complementarity constraint
\[
0 \leq \mysub{ y(t) - \lb{y} }{\sigma_{l,1}} \perp 
\mysub{ \ub{y} - y(t) }{\sigma_{l,2}} \geq 0.
\]

\begin{figure}
    \centering
    \includegraphics[width=0.5\textwidth]{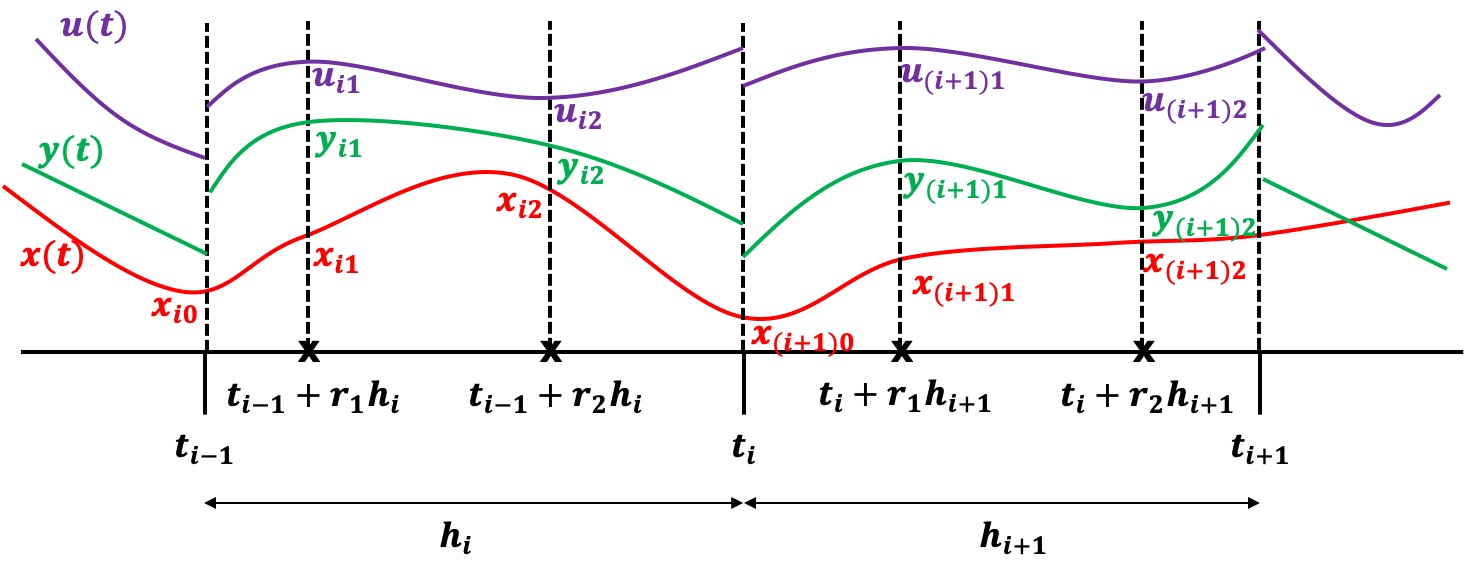}
    \caption{Collocation on finite elements with $\ncoll = 2$. Note that the differential variables ($x(t)$) is continuous across finite elements while the algebraic ($y(t)$) and control ($u(t)$) variables are not continuous across the finite elements. The Lagrange polynomial representing $x(t)$ on finite element $i$ is of degree $\ncoll$ and is parameterized by $x_{i0}, x_{i1}, x_{i2}$.  The Lagrange polynomial representing $y(t) (u(t))$ on finite element $i$ is of degree $(\ncoll-1)$ and is parameterized by $y_{i1}, y_{i2} (u_{i1}, u_{i2})$.}\label{fig:collocation}
\end{figure}

The dynamic optimization problem in~\eqref{dynopt} is transcripted to a NonLinear Program (NLP) by orthogonal collocation on finite elements.  The time interval $[t_0,t_f]$ is discretized into $\nfe$ finite elements of width $h_i$ such that $\sum_{i = 1}^{\nfe}h_i = t_f - t_0$. Let $t_i = t_0 + \sum_{i' \leq i} h_{i'}$ denote the ending time of the finite elements $i$.  The differential, algebraic and control variables in each finite element $i \in \{1,\ldots,\nfe\}$ are represented as Lagrange polynomials of degree $(\ncoll+1)$, $\ncoll$ and $\ncoll$ respectively where $\ncoll$ is the order of the collocation. Given $r_j \in (0,1]$ for $j = 1,\ldots,\ncoll$ the differential, algebraic and control variables in the $i$-th finite element, i.e. $t \in  [t_{i-1},t_i]$, are represented as the following polynomials 
\begin{subequations}
\begin{align}
    & x(t_{i-1} + \tau h_i) = \sum_{j=0}^{\ncoll} x_{ij} \Omega_j (\tau) \label{poly:x}\\
    & y(t_{i-1} + \tau h_i) = \sum_{j=1}^{\ncoll} y_{ij} \Psi_j (\tau) \label{poly:y} \\
    & u(t_{i-1} + \tau h_i) = \sum_{j=1}^{\ncoll} u_{ij} \Psi_j (\tau) \label{poly:u} \\
    & \text{ with } \Omega_j(\tau) = \prod_{k =0,\neq j}^{\ncoll} \frac{(\tau - r_k)}{(r_j - r_k)},\, 
    \Psi_j(\tau) = \prod_{k =1,\neq j}^{\ncoll} \frac{(\tau - r_k)}{(r_j - r_k)} \nonumber
\end{align}\label{poly}
\end{subequations}
where $r_0 = 0$, $\tau \in [0,1]$ and $x_{ij}, y_{ij}, u_{ij}$ denote the values of the differential, algebraic and control variables at time $(t_{i-i}+r_j h_i$). The values of $r_j$ are chosen as the roots of Legendre or shifted Radau polynomials of order $\ncoll$.  Figure~\ref{fig:collocation} provides a illustration of the polynomial representation of the differential, algebraic and control variables over the finite elements. 

\pyrobocop\, allows the user to employ a collocation method of order up to $\ncoll = 5$.  The user has the option of using either Legendre or shifted Radau roots.  In addition, the package also implements an explicit Euler scheme that can be useful in specifying discrete nonlinear systems as opposed DAEs.  Such discrete nonlinear systems may arise from learned machine learning models from data, such as Gaussian Process (GP) and Neural Network models. 

The NLP that results from applying the orthogonal collocation on finite elements to the dynamic optimization problem~\eqref{dynopt} is
\begin{subequations}
\begin{align}
    \min        &\, \sum\limits_{i=1}^{\nfe} h_i c(x_{ij},y_{ij},u_{ij}) \widehat{\Psi}_j + \phi(x_f,p) \label{nlp:obj} \\
    \text{s.t.} &\, f(\dot{x}_{ij},x_{ij},y_{ij},u_{ij}) = 0,\, x_{10} = x_0 \label{nlp:daeqn} \\
                &\, (\mysub{y_{ij}}{\sigma_{l,1}} - \nu_{l,1}) (\mysub{y_{ij}}{\sigma_{l,2}} - \nu_{l,2}) = 0 \,\forall\, l \in \setcompl \label{nlp:ceqn} \\
                &\, \lb{x} \leq x_{ij} \leq \ub{x}, \lb{y} \leq y_{ij} \leq \ub{y}, \lb{u} \leq u_{ij} \leq \ub{u} \label{nlp:bnds} \\
                &\, \dot{x}_{ij} = h_i \sum\limits_{k=0}^{\ncoll} x_{ik} \Omega'(\tau) \label{nlp:xdot} \\
                &\, x_{(i+1)0} = \sum\limits_{j=0}^{\ncoll} x_{ij} \Omega_j(1) \label{nlp:xcont} \\
                &\, x_f = \sum\limits_{j=0}^{\ncoll} x_{\nfe j} \Omega_j(1) \label{nlp:xf}
\end{align}\label{nlp}
\end{subequations}
where the decision variables in the~\eqref{nlp} are $x_{ij}$, $\dot{x}_{ij}$, $y_{ij}$ and $u_{ij}$. The variables $x_{ij}$ are indexed over $i \in \setnfe (:= \{1,\ldots,\nfe\})$ and $j \in \{0\} \cup \setncoll (:= \{1,\ldots,\ncoll\})$, while the rest of the variables are indexed over $i \in \setnfe$, $j \in \setncoll$. The constraints~\eqref{nlp:daeqn}-\eqref{nlp:xdot} are imposed for $i \in \setnfe$, $j \in \setncoll$. The constraint in~\eqref{nlp:xcont} imposes the continuity of the differential variable across the finite elements and is imposed for $i \in \setnfe \setminus \{\nfe\}$. The constraint~\eqref{nlp:xf} defines the state at the final time. 
The polynomial representation of $x$ within a finite element $i$ is used to relate $\dot{x}_{ij}$ to $x_{ij}$ in~\eqref{nlp:xdot} where $\Omega'(\tau)$ is the derivative of $\Omega(\tau)$ w.r.t. $\tau$. The notation $\widehat{\Psi}_j = \int_0^1 \Psi_j(\tau) d\tau$. If complementarity constraints are present then~\eqref{nlp} is an instance of a Mathematical Program with Complementarity Constraints (MPCC).

MPCCs are well known to fail the standard Constraint Qualification (CQ) such as the Mangasarian Fromovitz CQ (MFCQ), see \cite{luo1996mathematical}.  Hence, solution of MPCCs has warranted careful handling of the complementarity constraints when used in Interior Point Methods for NLP (IPM-NLP) using relaxation~\citep{ipoptc,twosidedipm} or penalty formulations~\citep{knitroc}.  In the case of active set methods, the robust solution of MPCC relies on special mechanism such as the elastic mode \citep{elasticmode}.  

\pyrobocop\, implements two possible relaxation schemes for complementarity constraints 
\begin{subequations}
\begin{align}
    \alpha_l (\mysub{y_{ij}}{\sigma_{l,1}} - \nu_{l,1}) (\mysub{y_{ij}}{\sigma_{l,2}} - \nu_{l,2}) \leq \delta \,\forall\, l \in \setcompl \label{ceqn:opt1} \\
   \sum\limits_{l \in \setcompl} \sum\limits_{j =1}^{\ncoll} \alpha_l (\mysub{y_{ij}}{\sigma_{l,1}} - \nu_{l,1}) (\mysub{y_{ij}}{\sigma_{l,2}} - \nu_{l,2}) \leq \delta n_c \label{ceqn:opt2}
\end{align}
\end{subequations}
where $\alpha_l = 1$ if the involved bounds ($\nu_{l,1},\nu_{l,2}$) are either both lower or both upper bounds. If one of the bounds is a lower bound and other is an upper bound then $\alpha_l$ is set to $-1$. Note that the choice of $\alpha_l$ ensures that the resulting product is nonnegative whenever~\eqref{nlp:bnds} are satisfied.  The first approach relaxes each complementarity constraint by a positive parameter $\delta$~\citep{ipoptc} while the second approach imposes the relaxation on the summation of all the complementarity constraints over a finite element $i$~\citep{elasticmode}.  In addition, we also have flexibility to keep the $\delta$ fixed to a constant parameter through out the optimization or link this with the barrier parameter in IPM-NLP~\citep{ipoptc,twosidedipm}.

When using time-stepping methods in the presence of complementarity constraints a lower integration error can only be obtained if the discontinuity is resolved accurately. Time-stepping methods for complementarity systems~\citep{PangStewart2008} do not employ a discontinuity resolution scheme and hence, the use of higher order collocation method is not meaningful.  The approach in \cite{biegler2019} addresses this problem by enforcing that the contacts do not change within a finite element. This allows them to employ higher order collocation. However, the resulting optimization problem tends to be more challenging and requires careful initialization for convergence. For the systems with complementarity constraints~\eqref{dynopt:ceqn}, the current version of the package only allows the choice of $\ncoll = 1$ which corresponds to explicit or implicit Euler scheme.

\section{Trajectory Optimization Problems in Robotics}\label{sec:prob_instants}
% [\devesh{This looks a little abrupt. Also, we should replace figure 2. Also (10) becomes explicit Euler when in the last section we explain collocation-based integration. Reviewers will complain about it.}] \\
\begin{figure}
    \centering
    \includegraphics[scale=0.35]{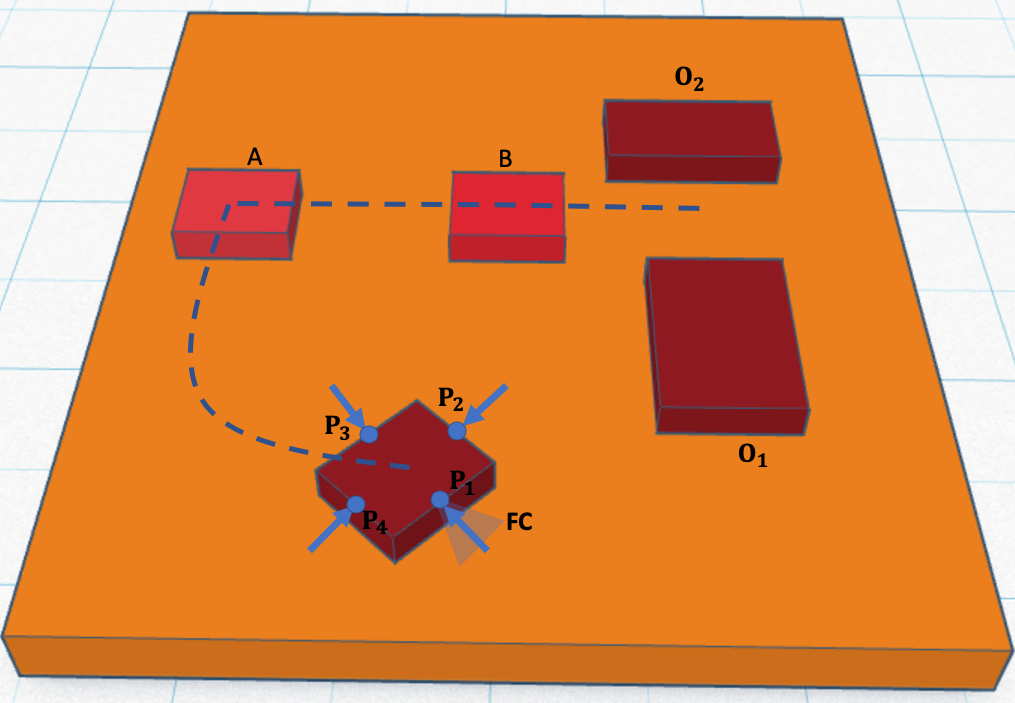}
    \caption{Planar pushing in the presence of obstacles. Our proposed formulation in \pyrobocop\ allows us to solve the collision avoidance with hybrid dynamics simultaneously using a novel formulation. The figure shows a possible solution to the given planning scenario. The figure shows a scenario where there are four possible points of contact for a pusher denoted as $P_i$, $i=1,\dots,4$. An approximate friction cone is also shown at the point of contact.}
    \label{fig:pushing}
\end{figure}
In this section, we describe three main instantiations of the mathematical programs that could be solved via the formulation described in Section~\ref{sec:probdesc}. These are problems with hybrid dynamics (e.g., systems with contacts using complementarity and mode enumeration) and problems with non-differentiable constraints like collision avoidance. It is noted that trajectory optimization problems for systems with smooth constraints are a subset of these problems and thus, are naturally solved by \pyrobocop.

\subsection{Manipulation Problems with Contacts}\label{subsec:manipulation_problem}

Robotic manipulation utilizes contacts to manipulate objects in its environment. Contacts appear in a lot of tasks in dexterous manipulation like pushing~\citep{LM96, doshi2020hybrid, zhang2020cazsl}, prehensile manipulation~\citep{chavan2020planar}, etc. A common manipulation primitive that we come across in robotics is pushing a body on a flat surface. Complementarity conditions could be used to model Coulomb friction between the pusher and the slider. Similarly, complementarity conditions could also appear because the pusher can exert force on a certain face of the slider (the object being pushed) only on contact with that face, and thus the distance function between the contact surface and the pusher will be zero depending on which surface the contact happens (see Figure~\ref{fig:pushing}). 

\begin{figure}
    \centering
    \includegraphics[scale=0.65]{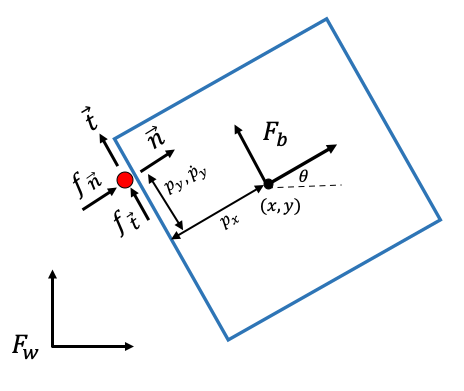}
    \caption{A schematic of a planar pusher-slider system. State of the system is given by $[x,y,\theta,p_y]^T$ assuming that the pusher only comes in contact with the left edge as shown in the figure. The world frame and the body frame of reference are denoted by $F_w$ and $F_b$ respectively.}
    \label{fig:pushing_analytical}
\end{figure}

For sake of exposition, we will briefly describe a planar pushing model. For more detailed description of the pushing model, readers are referred to~\cite{hogan2018reactive} and \cite{pmlr-v87-bauza18a}. A schematic for a pusher-slider system is shown in Figure~\ref{fig:pushing_analytical}. The frictional interaction between the pusher and slider leads to a linear complementarity system which we describe next. The pusher interacts with the slider by exerting forces in the normal and tangential direction denoted by $f_{\overrightarrow{n}}$, $f_{\overrightarrow{t}}$ (as shown in Figure~\ref{fig:pushing_analytical}) as well as a torque $\tau$ about the center of the mass of the object. Assuming quasi-static interaction, the limit surface~\citep{GOYAL1991307} defines an invertible relationship between applied wrench $\mathbf{w}$ and the twist of the slider $\mathbf{t}$. The applied wrench $\mathbf{w}$ causes the object to move in a perpendicular direction to the limit surface $\mathbf{H(w)}$. Consequently, the object twist in body frame is given by $\mathbf{t} = \nabla \mathbf{H(w)}$, where the applied wrench $\mathbf{w} = [f_{\overrightarrow{n}},f_{\overrightarrow{t}},\tau]$ could be written as $\mathbf{w} = \mathbf{J}^T(\overrightarrow{n} f_{\overrightarrow{n}} +\overrightarrow{t}f_{\overrightarrow{t}})$. For the contact configuration shown in Figure~\ref{fig:pushing_analytical}, the normal and tangential unit vectors are given by $\overrightarrow{n}=[1\quad 0]^T$ and $\overrightarrow{t}=[0\quad 1]^T$. The Jacobian $\mathbf{J} $ is given by $\mathbf{J}=$ $
\begin{bmatrix}
1 & 0 & -p_y \\
0 & 1 & p_x
\end{bmatrix}$.

The equation of motion of the pusher-slider system is then given by 
\begin{equation}
    \mathbf{\dot{x}}= \mathbf{f(\mathbf{x},\mathbf{u})}= \begin{bmatrix}
\mathbf{Rt}\\
\dot{p}_y
\end{bmatrix}
\label{eqn:pushing_dynamics}
\end{equation}
where $\mathbf{R}$ is the rotation matrix given by $\mathbf{R}=$
$\begin{bmatrix}
\cos\theta & -\sin\theta & 0\\
\cos\theta & \sin\theta & 0\\
0 & 0 & 1
\end{bmatrix}.$ Since the wrench applied on the system depends of the point of contact of pusher and slider, the state of the system is given by $\mathbf{x}= [x\quad y\quad \theta \quad p_y]^T$ and the input is given by $\mathbf{u} = [f_{\overrightarrow{n}}\quad f_{\overrightarrow{t}}\quad \dot{p}_y]^T$. The elements of the input vector must follow the laws of coulomb friction which can be expressed as complementarity conditions as follows:
\begin{equation}
 0 \leq   \dot{p}_{y+} \perp (\mu_p f_{\overrightarrow{n}}(t)-f_{\overrightarrow{t}}(t)) \geq 0 \nonumber \\
\end{equation}
\begin{equation}
 0 \leq   \dot{p}_{y-} \perp (\mu_p f_{\overrightarrow{n}}(t)+f_{\overrightarrow{t}}(t)) \geq 0  
 \label{eqn:compl_pushing}
\end{equation}

where $\dot{p}_y=\dot{p}_{y+}-\dot{p}_{y-}$ and the $\mu_p$ is the coefficient of friction between the pusher and the slider. The complementarity conditions in Eq.~\eqref{eqn:compl_pushing} mean that both $\dot{p}_{y+}$ and $\dot{p}_{y-}$ are non-negative and only one of them is non-zero at any time instant. Furthermore, $\dot{p}_y$ is non-zero only at the boundary of friction-cone. Consequently, the slipping velocity $\dot{p}_y$ cannot be chosen as an independent control input and is optimized while satisfying the conditions in Eq.~\eqref{eqn:compl_pushing}.

% As a result of these complementarity conditions in Eq.~\eqref{eqn:compl_pushing}, all the three elements can not be independently provided as inputs, and thus, $\dot{p}_y$ comes out of the complementarity conditions presented above.

% \devesh{Can you a describe contact model leading to complementarity constraints here?}

\subsection{Collision Avoidance}\label{subsec:collision_avoidance}
In this section, we present a novel formulation for the collision avoidance problem of rigid bodies. 
Path constraints arise in robotic systems due 
to their operation in cluttered environments.  The constraints 
impose that the trajectories of robots do not collide with other obstacles which may be static or dynamic. For example, the pushing scenario shown in Figure~\ref{fig:pushing} shows a frictional interaction scenario in the presence of obstacles. The goal in the example shown in Figure~\ref{fig:pushing} is to move the object being pushed between the two obstacles $O_1$ and $O_2$.  

%\devesh{Re-phrase to be consistent with rest of the draft.}\\
We propose a novel formulation for collision avoidance using complementarity constraints. The distinguishing features of the formulation are that: (i) it is differentiable and allows for the use of NLP solvers and (ii) the treatment is identical for the case of robot-static obstacle and robot-robot collision avoidance.  

\begin{figure}
    \centering
    \includegraphics[scale=0.45]{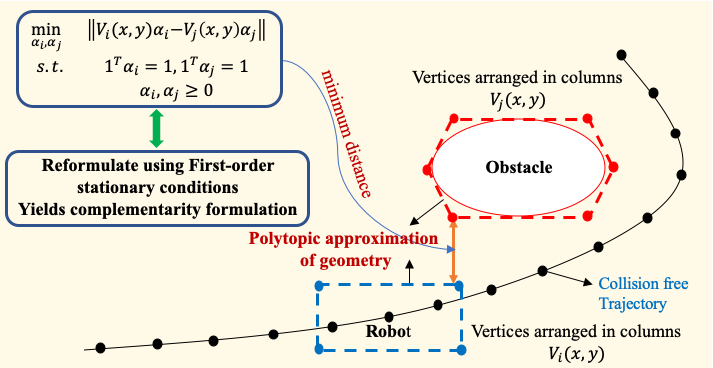}
    \caption{Schematic showing the approximations needed to formulate an MPCC for the collision avoidance problem in \pyrobocop. Using a polytopic representation for the objects in the environment allows us to represent first-order stationary conditions for minimum distance function as complementarity constraints.}
    \label{fig:collision_avoidance_analytical}
\end{figure}

We assume that the extent of the robot and obstacles are modeled as polytopes in 3-D and are specified by the set of vertices (see Figure~\ref{fig:collision_avoidance_analytical}).  Let $n_O$ denote the number of objects (including the robot and the obstacles).  The polytope bounding the objects at time $t$ are denoted by ${\cal O}_i(t)$ for $i = 1,\ldots,n_O$.  We assume that user provides the matrix $V_i(x(t),y(t)) \in \R^{3 \times n_{vi}}$ with columns representing the coordinates of the $n_{vi}$ vertices of the polytope ${\cal O}_i(t)$.  The polytope ${\cal O}_i(t)$ is 
\begin{align}
    {\cal O}_i(t) = \{ q \,|\, q = V_i(x(t),y(t))\alpha, 1_{n_{vi}}^T\alpha = 1, \alpha \geq 0 \} \label{defobject}
\end{align}
where $1_{n_{vi}} \in \R^{n_{vi}}$ is a vector of all ones and $\alpha \in \R^{n_{vi}}$ are nonnegative vectors that generate the convex hull of the vertices.  
The dependence of the coordinates's vertices on the position, orientation of the object is modeled through the functional dependence of $V_i$ on $x(t),y(t)$. Note that a static obstacle's vertex coordinates are independent of $x(t),y(t)$. 

Consider two objects $i,j \in \{1,\ldots,n_{O}\}$. The distance between the objects at time $t$ is 
\begin{align}
    \min \|q_i - q_j\|^2 \text{ s.t. } q_i \in {\cal O}_i(t) , q_j \in {\cal O}_j(t). \label{defdist}
\end{align}
Suppose that the minimum separation required between the two objects is $\epsilon_{ij} > 0$. \cite{GilbertJohnson1985} proposed to model the distance constraint as
\begin{align}
    \|q_i - q_j\| \geq \epsilon_{ij} \text{ where } 
    q_i, q_j \text{ solves } \eqref{defdist}. \label{implicitdistform}
\end{align}
The formulation~\eqref{implicitdistform} is at best once differentiable when the minimizer of~\eqref{defdist} is unique.  The uniqueness requirement cannot be guaranteed when the objects are modeled as polytopes. As a consequence, the formulation in~\cite{GilbertJohnson1985} cannot be directly provided to NLP solvers.  Our formulation overcomes these drawbacks.  The price to pay is that we need to solve a MPCC instead of a NLP. A schematic representing the underlying idea is shown in Figure~\ref{fig:collision_avoidance_analytical}. 

The minimization problem~\eqref{defdist} can be equivalently posed as
\begin{subequations}
\begin{align}
    \min\limits_{\alpha_{ij,i},\alpha_{ij,j}} &\; \| V_i(x(t),y(t))\alpha_{ij,i} - V_j(x(t),y(t))\alpha_{ij,j} \|^2 \\
    \text{s.t.} &\; 1^T_{n_{vi}}\alpha_{ij,i} = 1, 1^T_{n_{vj}}\alpha_{ij,j} = 1,
    \alpha_{ij,i}, \alpha_{ij,j} \geq 0
\end{align}\label{defdistalpha}
\end{subequations}
where $\alpha_{ij,i}$ and $\alpha_{ij,j}$ are variables denoting the convex combinations of the vertices of the polytope bounding the objects $i$, $j$ (see Figure~\ref{fig:collision_avoidance_analytical}).  At an optimal solution $\alpha^*_{ij,i}, \alpha^*_{ij,j}$ to~\eqref{defdistalpha} the points $V_i(x(t),y(t))\alpha^*_{ij,i}$ and $V_j(x(t),y(t))\alpha^*_{ij,j}$ are respectively the points in objects $i,j$ that give the shortest distance between the objects. 
The optimization problem in~\eqref{defdistalpha} is a convex problem. Every first order stationary point of~\eqref{defdistalpha} is a minimizer~\citep{Boydbook}.  This suggests imposing~\eqref{defdistalpha} through its first order stationary conditions
\begin{subequations}
\begin{align}
    & V_i(x(t),y(t))^T(V_i(x(t),y(t))\alpha_{ij,i} - V_j(x(t),y(t))\alpha_{ij,j}) 
    \nonumber \\ 
    & + 1_{n_{vi}}\beta_i - \nu_i = 0 \label{diststats.c1} \\
    & V_j(x(t),y(t))^T(V_j(x(t),y(t))\alpha_{ij,j} - V_i(x(t),y(t))\alpha_{ij,i}) 
    \nonumber \\ 
    & + 1_{n_{vj}}\beta_j - \nu_j = 0 \label{diststats.c2} \\
    & 1_{n_{vi}}^T\alpha_{ij,i} = 1, 1_{n_{vj}}^T\alpha_{ij,j} = 1 \label{diststats.c3} \\
    & 0 \leq \mysub{\alpha_{ij,i}}{k} \perp \mysub{\nu_i}{k} \geq 0 \,\forall\, k = 1,\ldots,n_{vi} \label{diststats.c4} \\ 
    & 0 \leq \mysub{\alpha_{ij,i}}{k} \perp \mysub{\nu_i}{k} \geq 0 \,\forall\, k = 1,\ldots,n_{vj} \label{diststats.c5} 
\end{align}\label{diststats}
\end{subequations}
where $\beta_i, \beta_j$ are the multipliers for the equality constraints in~\eqref{diststats.c3} and $\nu_i, \nu_j$ are the multipliers for the nonnegative bounds on $\alpha_{ij,i}, \alpha_{ij,j}$. 
The separation requirement can be modeled as
\begin{align}
    & \sqrt{ \|V_i(x(t),y(t))\alpha_{ij,i} - V_j(x(t),y(t))\alpha_{ij,j} \|^2 + \epsilon^2} \nonumber \\
    & \geq \sqrt{\epsilon_{ij}^2 + \epsilon^2} \label{ourdist} 
\end{align}
The parameter $\epsilon > 0$ is a small constant that is included to render the constraint~\eqref{ourdist} differentiable everywhere.  The separation requirements between the $n_O$ objects are modeled through the constraints \eqref{diststats}-\eqref{ourdist} for all $i < j$, with $ i,j \in \{1,\ldots,n_O\}$. 

The variables $\alpha_{ij,i}, \beta_i, \nu_i, \alpha_{ij,j}, \beta_j, \nu_j$ are in fact trajectories over time i.e. the variables are different for every time instant at which the collision avoidance constraint is imposed.  The number of additional variables and constraints introduced for a pair of objects $i,j$ is $(2n_{vi}+2n_{vj}+2)$. On applying the discretization, the total number of additional variables and constraints added to the MPCC is $N_en_O(n_O-1)(n_{vi}+n_{vj}+1)$.  The size of the MPCC scales quadratically in the number of objects and linearly in the number of vertices defining the objects. The constraints modeling collision avoidance are sparse.  A sparsity preserving automatic differentiation algorithm and an NLP solver that can exploit such sparsity are critical to obtaining computational efficiency.  We will touch upon in the design of our software next.

\begin{figure*}[!h]
    \centering
    \includegraphics[scale=0.85]{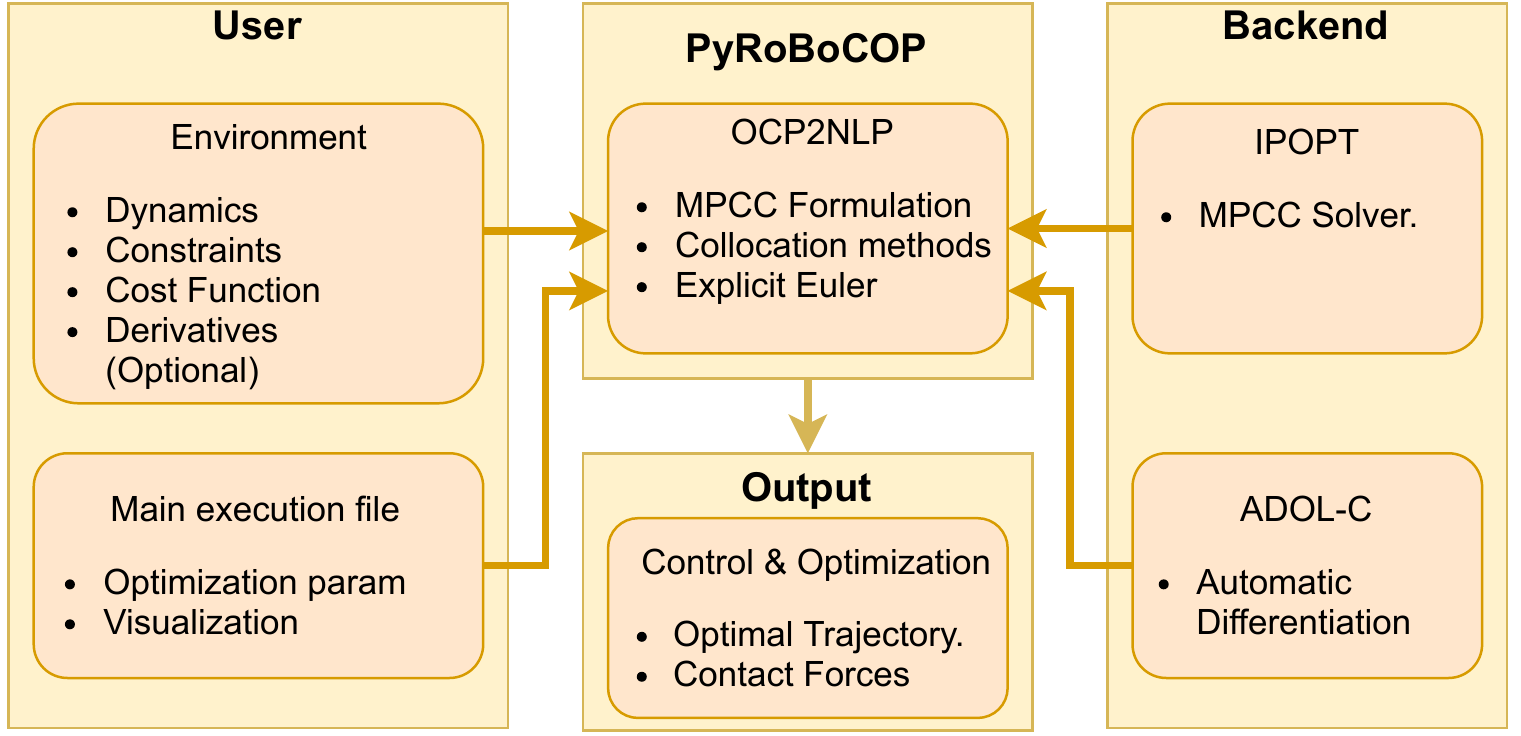}
    \caption{Workflow in \pyrobocop. The dynamics provided by the user to create a MPCC which is then optimized using \ipopt\ and the gradients are evaluated using automatic differentiation via \adolc. }
    \label{fig:pyrobocop}
\end{figure*}

\subsection{Optimizing Over Mode Sequences}\label{subsec:mode_sequences}

Consider the complementarity constraint in~\eqref{dynopt:ceqn}. A feasible point for~\eqref{dynopt:ceqn} requires either
\begin{align}
    \mysub{y(t)}{\sigma_{l,1}} - \nu_{l,1} = 0 \text{ or } \mysub{y(t)}{\sigma_{l,2}} - \nu_{l,2} =0 \label{ceqnasdisj}
\end{align}
for each $l \in \setcompl$. At every instant of time, the system can choose to enforce either one of the equalities in~\eqref{ceqnasdisj} for each $l \in \setcompl$. This requirement reveals that~\eqref{dynopt:ceqn} embeds a disjunctive structure. This disjunctive nature of the system can be fully exposed through the notion of \emph{modes}. At a particular time the mode of the system is defined as the set $m = \{(l,b(l)) \,|\, l \in \setcompl, b(l) \in \{1,2\}\}$ denoting which of the the two inequalities are satisfied as equality for each $l \in \setcompl$. Note that set of of all possible modes $\setmodes$ has cardinality $2^{|\setcompl|}$.  The formulation in~\eqref{dynopt} allows the system to be in any of modes in $\setmodes$ at each instant of time assuming that the rest of the constraints can be satisfied.  In certain robotic tasks, the sequence of modes is specified apriori but the time duration of the mode is not known (for example, consider a table-top manipulation scenario which can be performed using a known contact sequence, and thus known mode sequence).  We can utilize \pyrobocop\, to optimize the operation of such systems. We describe the continuous time-formulation for optimizing over mode sequences. The procedure described in~\ref{sec:probdesc} can be followed to convert the dynamic optimization problem to a nonlinear program.

Suppose for simplicity that the system is constrained to operate in two modes where the first mode is $m_1 = \{(l,1) \,|\, l \in \setcompl\}$ and the second mode is $m_2 = \{(l,2) \,|\, l \in \setcompl\}$.  As mentioned earlier, the time duration for each of the modes needs to be determined as part of the optimization.  Let $T_1,T_2$ denote the time duration for the two modes.  Note that these are now parameters for optimization.  Further, the duration of time in each mode is normalized to $1$ instead of $T_1,T_2$ respectively and we employ a scaled time $\tilde{t} \in [0,2]$.  The system is in mode $m_1$ for $\tilde{t} \in [0,1]$ and is in mode $m_2$ for $\tilde{t} \in [1,2]$.  The key advantage is that switching between modes can be precisely fixed in the scaled time ($\tilde{t} = 1$) coordinates which could not be done in absolute time coordinates.  With this scaling of time, the variable $\dot{x}$ is equal to $\frac{1}{T_1}\tilde{\dot{x}}$ for $\tau \in [0,1]$ and is equal to $\frac{1}{T_2}\tilde{\dot{x}}$ for $\tau \in [1,2]$.  The dynamics of the system~\eqref{dynopt:daeqn} can be recast as 
\begin{subequations}
    \begin{align}
        f\left(\frac{1}{T_1}\tilde{\dot{x}}(\tilde{t}),y(\tilde{t}),u(\tilde{t}),p\right) = 0 \,\forall\, \tilde{t} \in [0,1] \\
        f\left(\frac{1}{T_2}\tilde{\dot{x}}(\tilde{t}),y(\tilde{t}),u(\tilde{t}),p\right) = 0 \,\forall\, \tilde{t} \in (1,2]. 
    \end{align}\label{dynopt_modeseq_daeqn}
\end{subequations}
%
%
%
% FIGURE FROM SOFTWARE DESCTPTION SECTION

%
%
%
Further, the mode sequence is now realized by imposing time-varying bounds on the algebraic variables in the complementarity constraints, i.e. 
\begin{subequations}
    \begin{align}
        \mysub{y(\tilde{t})}{\sigma_{l,1}} \in 
        \left\{ \begin{aligned}
            \left[ \nu_{l,1}, \nu_{l,1} \right] \text{ for } \tilde{t} \in [0,1] \\
            [\mysub{\underline{y}}{\sigma_{l,1}},\mysub{\overline{y}}{\sigma_{l,1}}] \text{ for } \tilde{t} \in (1,2]
        \end{aligned}\right. \\
        \mysub{y(\tilde{t})}{\sigma_{l,2}} \in 
        \left\{ \begin{aligned}
            \left[\mysub{\underline{y}}{\sigma_{l,2}},\mysub{\overline{y}}{\sigma_{l,2}}\right] \text{ for } \tilde{t} \in [0,1] \\
            [\nu_{l,2},\nu_{l,2}] \text{ for } \tilde{t} \in (1,2] 
            \end{aligned}\right.
    \end{align}\label{dynopt_modeseq_bnds}
\end{subequations}
The dynamic optimization problem over the mode sequence can be  written as 
\begin{subequations}
\begin{align}
    \min\limits_{x,y,u,p}        &\, T_1\int\limits_{0}^{1} c(x(\tilde{t}),y(\tilde{t}),u(\tilde{t}),p) d\tilde{t} + \nonumber \\
                &\, T_2\int\limits_{1}^{2} c(x(\tilde{t}),y(\tilde{t}),u(\tilde{t}),p) d\tilde{t} + \phi(x(2),p)\label{dynopt_modeseq:obj} \\
    \text{s.t.} &\, \text{Eq. }\eqref{dynopt_modeseq_daeqn},\, x(0) = x_0 \label{dynopt_modeseq:daeqn} \\
                &\, \text{Eq. }\eqref{dynopt_modeseq_bnds} \,\forall\, l \in \setcompl \label{dynopt_modeseq:ceqn} \\
                &\, \lb{x} \leq x(\tilde{t}) \leq \ub{x}, \lb{y} \leq y(\tilde{t}) \leq \ub{y}, \lb{u} \leq u(\tilde{t}) \leq \ub{u} \label{dynopt_modeseq:bnds}
\end{align}\label{dynopt_modeseq}
\end{subequations}
The discretization can be applied to~\eqref{dynopt_modeseq} to obtain a nonlinear program. Note that this formulation does not have complementarity constraints.

% \begin{figure*}%[!h]
%     \centering
%     \includegraphics[scale=0.85]{figures/PyRoBoCOP.pdf}
%     \caption{Workflow in \pyrobocop. The dynamics provided by the user to create a MPCC which is then optimized using \ipopt\ and the gradients are evaluated using automatic differentiation via \adolc. }
%     \label{fig:pyrobocop}
% \end{figure*}

\section{Software Description}\label{sec:software_description}

% \begin{figure*}
%     \centering
%     \includegraphics[scale=0.55]{figures/pyrobocop_intro.png}
%     \caption{Workflow in \pyrobocop. The dynamics provided by the user to create a MPCC which is then optimized using \ipopt\ and the gradients are evaluated using automatic differentiation via \adolc. }
%     \label{fig:pyrobocop}
% \end{figure*}

%\diego{We should refer to the code repo and the guide that we wrote. But how?}
Figure~\ref{fig:pyrobocop} provides a high-level summary of the flow of control in \pyrobocop.  A user provided class specifies the dynamic optimization problem~\eqref{dynopt}. This is also briefly described in Figure~\ref{fig:pyrobocop}. The user needs to provide the equality constraints for the dynamical system. These constraints could include the dynamics information for the system, the bounds on the system state and inputs, and information about complementarity constraints, if any. Furthermore, a user needs to provide the objective function, and also has the option to provide derivative information (note the derivative information is optional).  \pyrobocop\, expects the user provided class to implement the following methods in order to formulate an MPCC (or NLP) (also shown in Figure~\ref{fig:pyrobocop}).
\begin{itemize}
    \item \texttt{get\_info}: Returns information on~\eqref{dynopt} including $n_d$, $n_a$, $n_u$, $|\setcompl|$, $n_p$, $N_e$, $h_i$.
    \item \texttt{bounds}: Returns the lower and upper bounds on the variables $x(t)$, $\dot{x}(t)$, $y(t)$, $u(t)$ at a time instant $t$.
    \item \texttt{initialcondition}: Returns the initial conditions for the variables $x(t_0)$, i.e. values of the differential variables at initial time instant $t_0$.
    \item \texttt{initialpoint}: Returns the initial guess for the variables $x(t)$, $\dot{x}(t)$, $y(t)$, $u(t)$ at a time instant $t$. This initial guess is passed to the NLP solver.
    \item \texttt{objective}: Implements method to evaluate and return  $c(x(t),y(t),u(t),p)$ at a time instant $t$.
    \item \texttt{constraint}: Implements method to evaluate and return $(f(x(t),\dot{x}(t),y(t),u(t),p)$ at a time instant $t$.
\end{itemize}
We provide a description of optional methods that are expected if certain specified conditions are satisfied.
\begin{itemize}
    \item \texttt{bounds\_finaltime}: Returns the bounds on the variables $x(t_f)$ at the final time. This method allows to specify a final time condition on a subset or all of the differential variables.
    \item \texttt{bounds\_params}: Returns information on lower and upper bounds on the parameters $p$. This method must be implemented if $n_p > 0$.
    \item \texttt{initialpoint\_params}: Returns the initial guess for the parameters $p$. This method  must be implemented if $n_p > 0$. This initial guess is passed the NLP solver.
    \item \texttt{get\_complementarity\_info}: Returns information on the complementarity constraints in~\eqref{dynopt} i.e. $\setcompl$ and also information on whether the lower or upper bound is involved in the complementarity constraint. This method must be implemented if $\setcompl \neq \emptyset$.
    \item \texttt{objective\_mayer}: Implements method to evaluate and return $\phi(x(t_f),p)$.
    \item \texttt{get\_objects\_info}: Returns the information on number of objects $n_O$, flags to indicate if these obstacles are static or dynamic and the number of vertices $n_{vi}$ for the polytope bounding the objects.
    \item \texttt{get\_object\_vertices}: Implements and returns the matrix $V_i(x(t),y(t)) \in \R^{3 \times n_{vi}}$ representing the vertices of the polytope bounding the objects. This method is called only when \texttt{get\_objects\_info} is implemented and $n_O > 0$.
\end{itemize}

%
% FIGURE PUSHING EXAMPLE
%
%
\begin{figure*}[!ht]
	\begin{minipage}[]{1.0\columnwidth}
		\centering
		\includegraphics[scale=0.5]{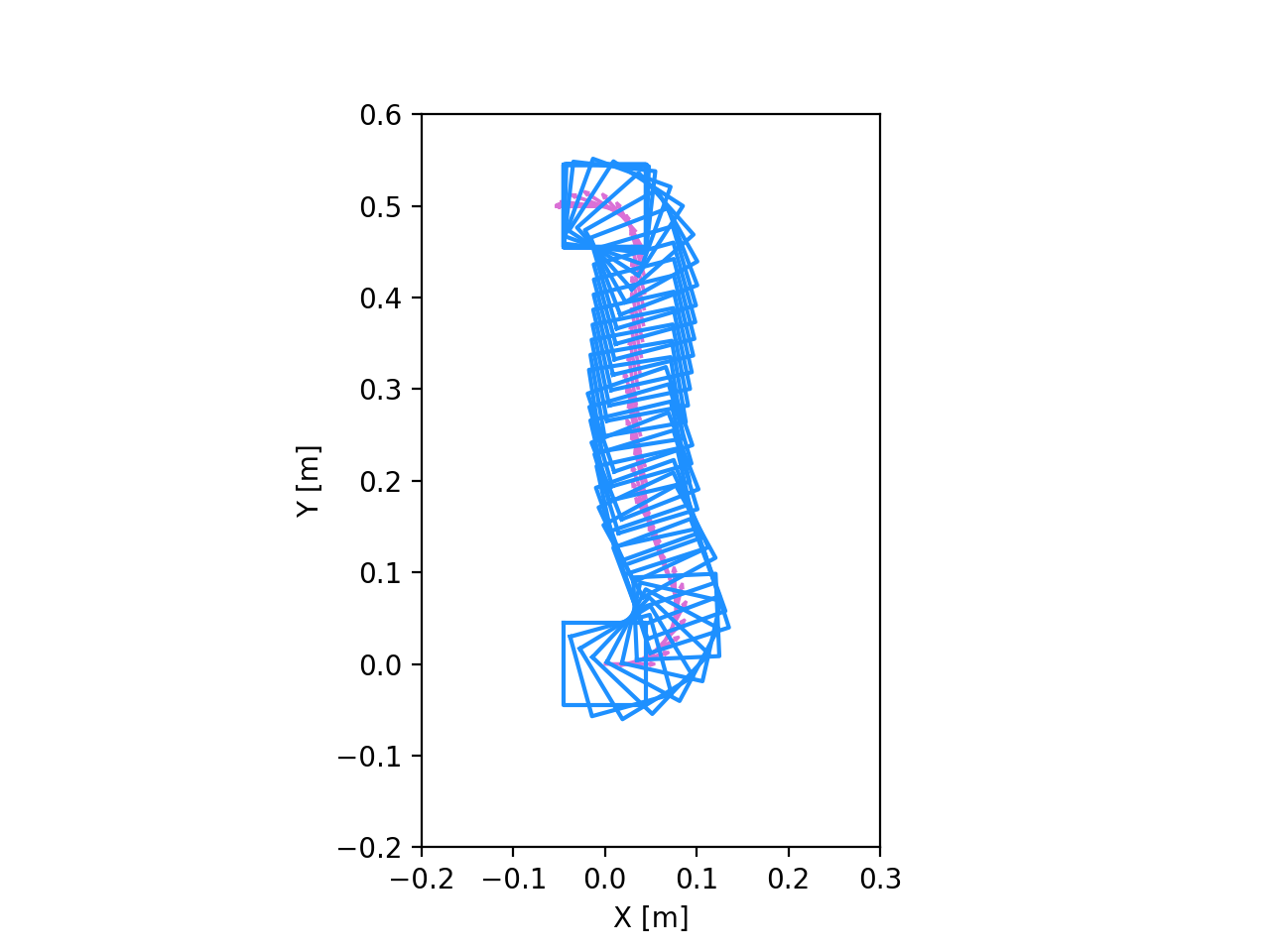}
		\vskip -0.05in
		\subcaption{Pushing Sequence for $ \mathbf{x_g}=(0,0.5,\pi)$}\label{fig:pushing_ex1}
	\end{minipage}
	\begin{minipage}[]{1.0\columnwidth}
		\centering
		\includegraphics[scale=0.5]{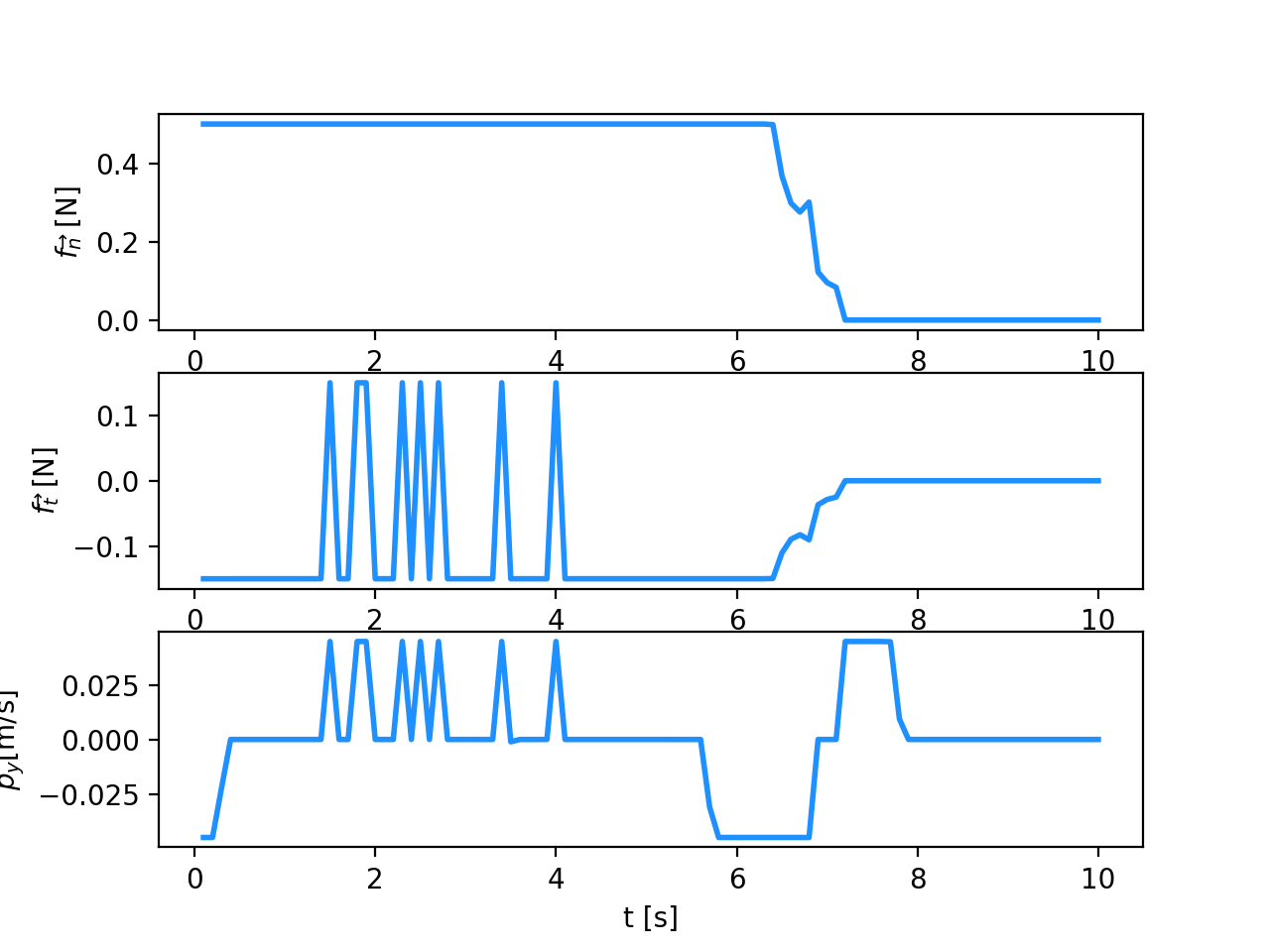}
		\vskip -0.05in
		\subcaption{Optimal Controls}\label{fig:pushing_ex1_input}
	\end{minipage}
	\begin{minipage}[]{1.0\columnwidth}
		\centering
		\includegraphics[scale=0.5]{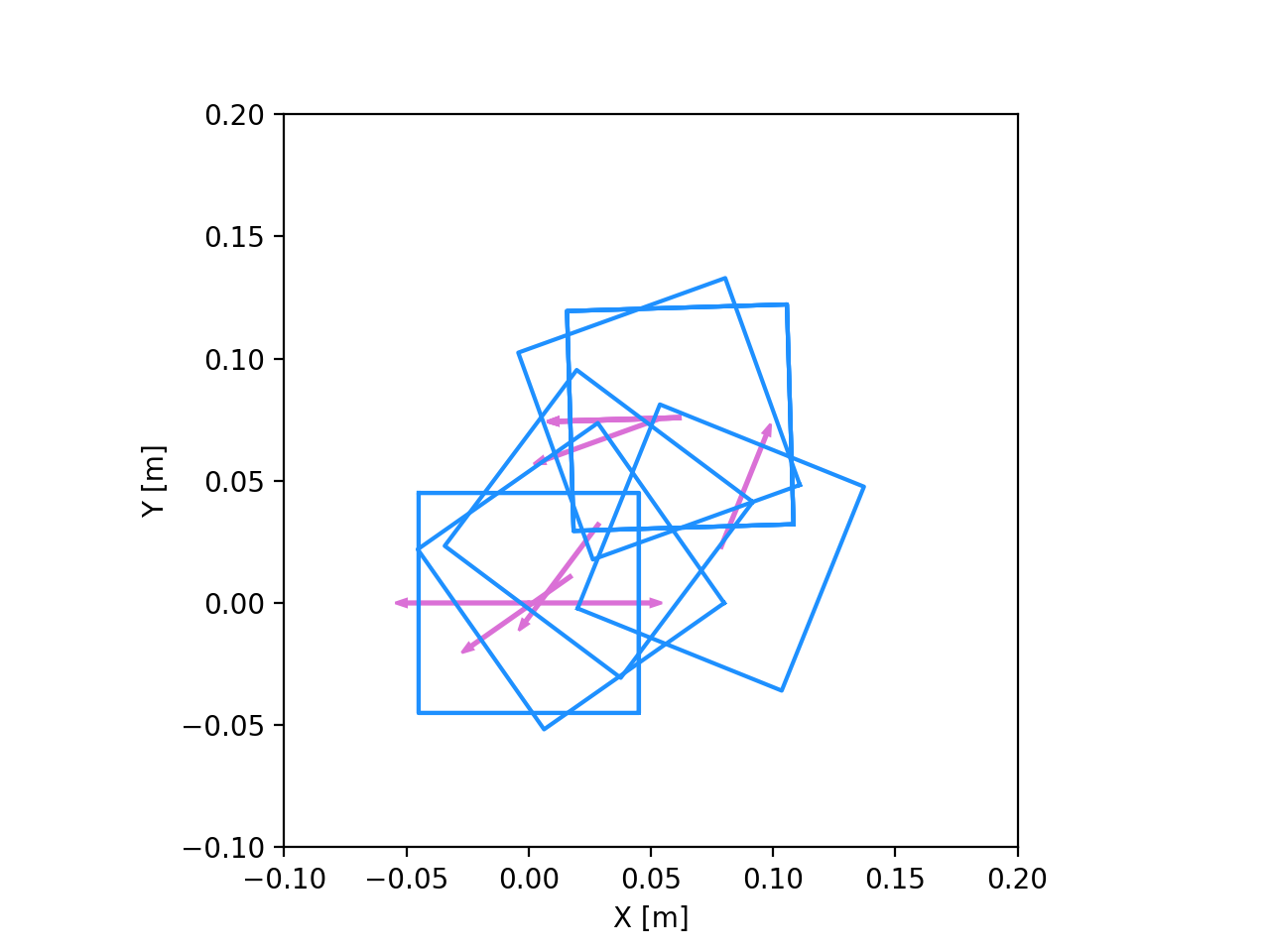}
		\vskip -0.05in
		\subcaption{Pushing sequence for $\mathbf{x_g}=(0,0,\pi)$}\label{fig:pushing_ex2}
	\end{minipage}
	\begin{minipage}[]{1.0\columnwidth}
		\centering
		\includegraphics[scale=0.5]{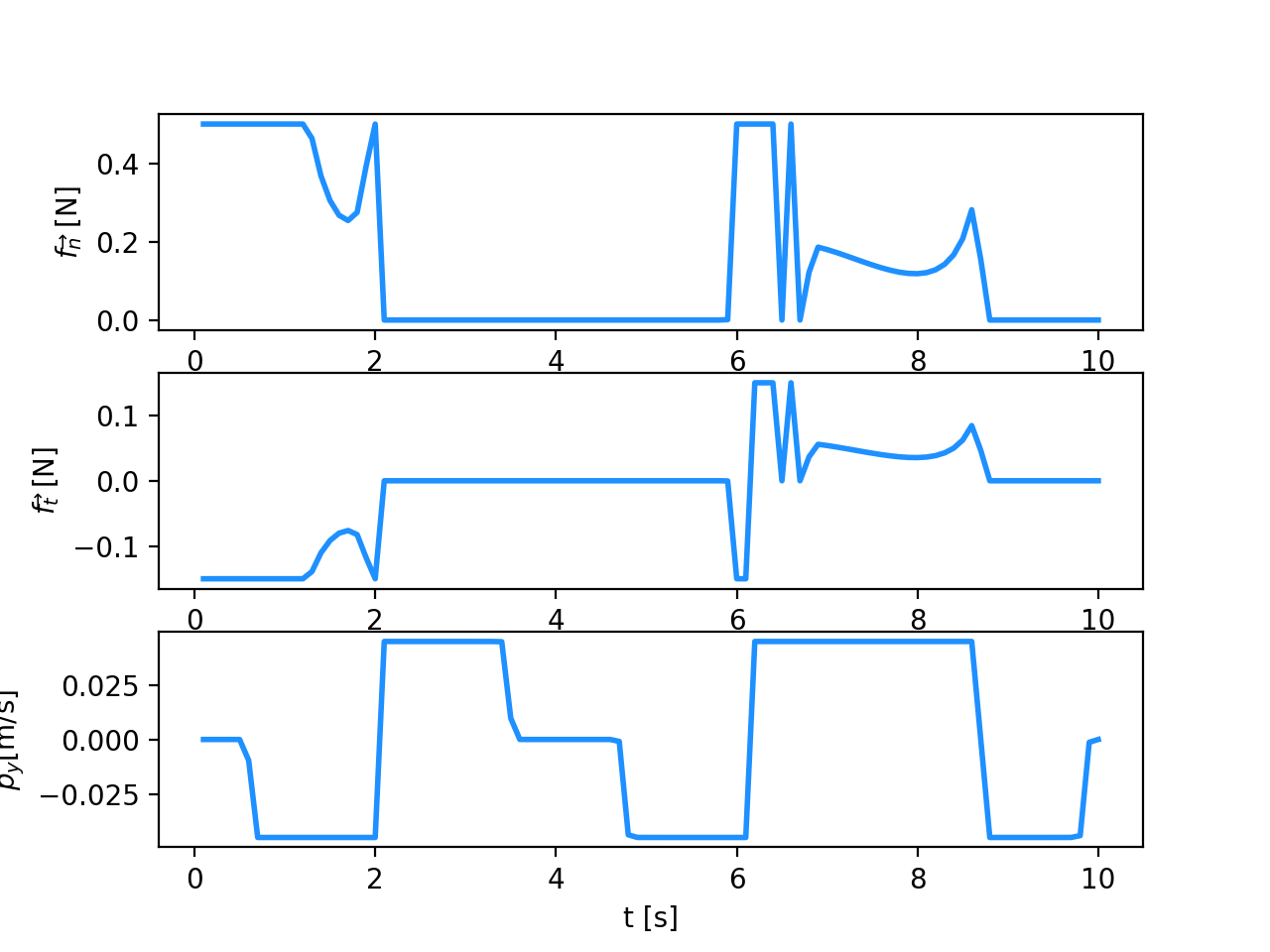}
		\vskip -0.05in
		\subcaption{Optimal Controls}\label{fig:pushing_ex2_input}
	\end{minipage}
	\vskip -0.05in
    \caption{Optimal pushing sequences and control inputs obtained by solving the MPCC for two different goal conditions. The switching sequence between sticking and slipping contact formation could be visualized by the trajectory of $\dot{p}_y$. The pusher maintains a sticking contact with the slider when $\dot{p}_y=0$. For clarity, we show very few frames of the pushing sequence in the second example in plot~\ref{fig:pushing_ex2}.}
	\label{fig:example_pushing}
\end{figure*}
Note that the user is not required to implement the collision avoidance constraints~\eqref{diststats}-\eqref{ourdist}. The user only provides the matrix $V_i(x(t),y(t))$ which models the dependence of the vertices of the bounding polytope on the differential and algebraic variables in~\eqref{dynopt}.  \pyrobocop\, defines a new class that wraps around the user-provided class to provide a dynamic optimization problem that is of the form in~\eqref{dynopt}.  The new class includes additional variables and constraints modeling the collision avoidance constraints.

\pyrobocop\, is interfaced with \adolc\,~\citep{ADOLC} to compute derivatives (see the Backend block in Figure~\ref{fig:pyrobocop}).  Note that the \adolc\, can also provide the sparsity pattern of the constraint jacobian and hessian of the Lagrangian. As mentioned earlier, the exploitation of sparsity in computations of the NLP is critical to solve large problems. To provide derivatives \pyrobocop\, used \adolc\, to set up tapes~\citep{ADOLC} for evaluating: (i) the objective~\eqref{nlp:obj}, (ii) constraints including the DAE~\eqref{nlp:daeqn} and a reformulation of~\eqref{nlp:ceqn}, and (iii) the Hessian of the Lagrangian of the NLP~\eqref{nlp}.  The set-up of the tape is done prior to passing control to the NLP solver.  The advantage of this approach is that evaluation of~\eqref{nlp:obj},~\eqref{nlp:daeqn} are now C-function calls instead of Python-function calls.  This considerably reduced the time spent in function evaluations for the NLP solver. As shown in Figure~\ref{fig:pyrobocop}, \pyrobocop\ uses \ipopt\ as the optimization solver. 

The user has flexibility in specifying how the complementarity constraints are solved.  The choices are: (i)~\eqref{ceqn:opt1} with $\delta$ fixed, (ii)~\eqref{ceqn:opt2} with $\delta$ fixed, (iii)~\eqref{ceqn:opt1} with $\delta$ set equal to the interior point barrier parameter, (iv)~\eqref{ceqn:opt2} with $\delta$ set equal to the interior point barrier parameter, and (v) the objective function is appended with complementarity terms as $\sum_{i=1}^{N_e}\sum\limits_{l \in \setcompl} \sum\limits_{j =1}^{\ncoll} \alpha_l (\mysub{y_{ij}}{\sigma_{l,1}} - \nu_{l,1}) (\mysub{y_{ij}}{\sigma_{l,2}} - \nu_{l,2})$.  The convergence behavior of formulations can be quite different and we provide these implementations so the user can choose one that works best for the problem at hand. A link to download and install \pyrobocop\ will be provided in the camera-ready version of the paper.
%In this section, we briefly describe the use of \pyrobocop.  This section is supposed to serve as a guide to help new users get familiar with \pyrobocop.  We explain how \pyrobocop\ can be used for new problems, and some parameters that user should be aware of while using it to solve their own problems.

\section{Numerical results}\label{sec:results}
In this section, we test \pyrobocop\ in several robotic simulations providing solutions to trajectory optimization problems including several systems with complementarity constraints. In all these examples, we do not provide a feasible initialization, and the performance of \pyrobocop\ would be enhanced with a better initialization.

To foster reproducibility, we make the source code for each of the following examples in the additional material.
% ADD LINK!!

\subsection{Planar Pushing}

In this section, we show some results for planar pushing without any obstacles. The model for planar pushing was earlier presented in Section~\ref{subsec:manipulation_problem} (see Eqs~\eqref{eqn:pushing_dynamics} and~\eqref{eqn:compl_pushing}). The complementarity constraints are used to represent slipping or sticking contact between the slider and the pusher. Two pushing trajectories with different goal configurations from the same initial state are shown in Figure~\ref{fig:example_pushing}. In both these examples, the initial pose of the slider is $\mathbf{x}_{\texttt{init}}=(0,0,0)$ and the desired goal pose of the slider is $\mathbf{x_g}=(0,0.5,\pi)$ and $(0,0,\pi)$. The initial point of contact between the pusher and the slider is $p_y=0$. For all these examples, the maximum normal force is set to $0.5 $ N and the coefficient of friction is $\mu_p=0.3$. The corresponding control trajectory shows the sequence of forces $f_n$ and $f_t$ used by the slider to obtain the desired trajectory. The plot of $\dot{p}_y$ shows the sequence of sticking and slipping contact as found by \pyrobocop\ and thus this also decides the contact point between the pusher and the slider. Note that the pusher maintains sticking contact with slider whenever $\dot{p}_y=0$, and slipping contact otherwise. In both these examples, the objective function is a function of the target state and the control inputs.

\subsection{Car Parking Example with Obstacle Avoidance}

\begin{figure}[!h]
    \centering
    \includegraphics[scale=0.45]{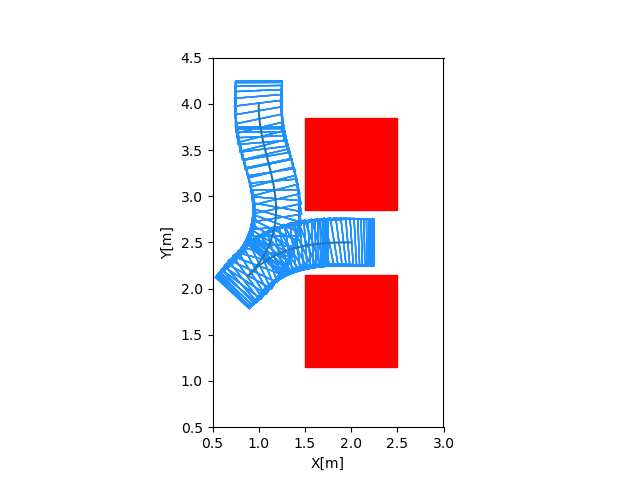}
    \caption{Motion Planning in the presence of obstacles using the proposed obstacle avoidance method using~complementarity constraints.}
    \label{fig:car_parking_problem}
\end{figure}

To show collision avoidance, we show a parking scenario which has been previously used to show the effectiveness of several optimization-based collision-avoidance methods~\citep{zhang2018autonomous}. The dynamics of the car is described in~\citep{zhang2018autonomous}. The state of the car is defined by a $4$-dimensional vector $\mathbf{x}=[x,y,\theta,v]^T$, where $x,y$ is the center of the rear axis, $\theta$ is the heading angle and $v$ is the longitudinal velocity of the car. The initial state of the car was chosen to be $\mathbf{x}_\texttt{init}=(1,4,0,0)$ and the desired state was chosen to be $\mathbf{x}_g=(2,2.5,\pi/2,0)$. The resulting optimal solution from \pyrobocop\ is shown in Figure~\ref{fig:car_parking_problem}. As described earlier in Section~\ref{subsec:collision_avoidance}, the two static obstacles are specified by providing the vertex set for them.

\subsection{Assembly of Belt Drive Unit}
An example of a complex manipulation problem that involves contacts, elastic objects and collision avoidance is provided by the Belt Drive Unit system. This assembly challenge was presented as one of the most challenging competition in the World Robot Summit 2018\footnote{\url{https://worldrobotsummit.org/en/about/}}~\citep{drigalski2019}. 
\begin{figure}[!h]
    \centering
    \includegraphics[width=\columnwidth]{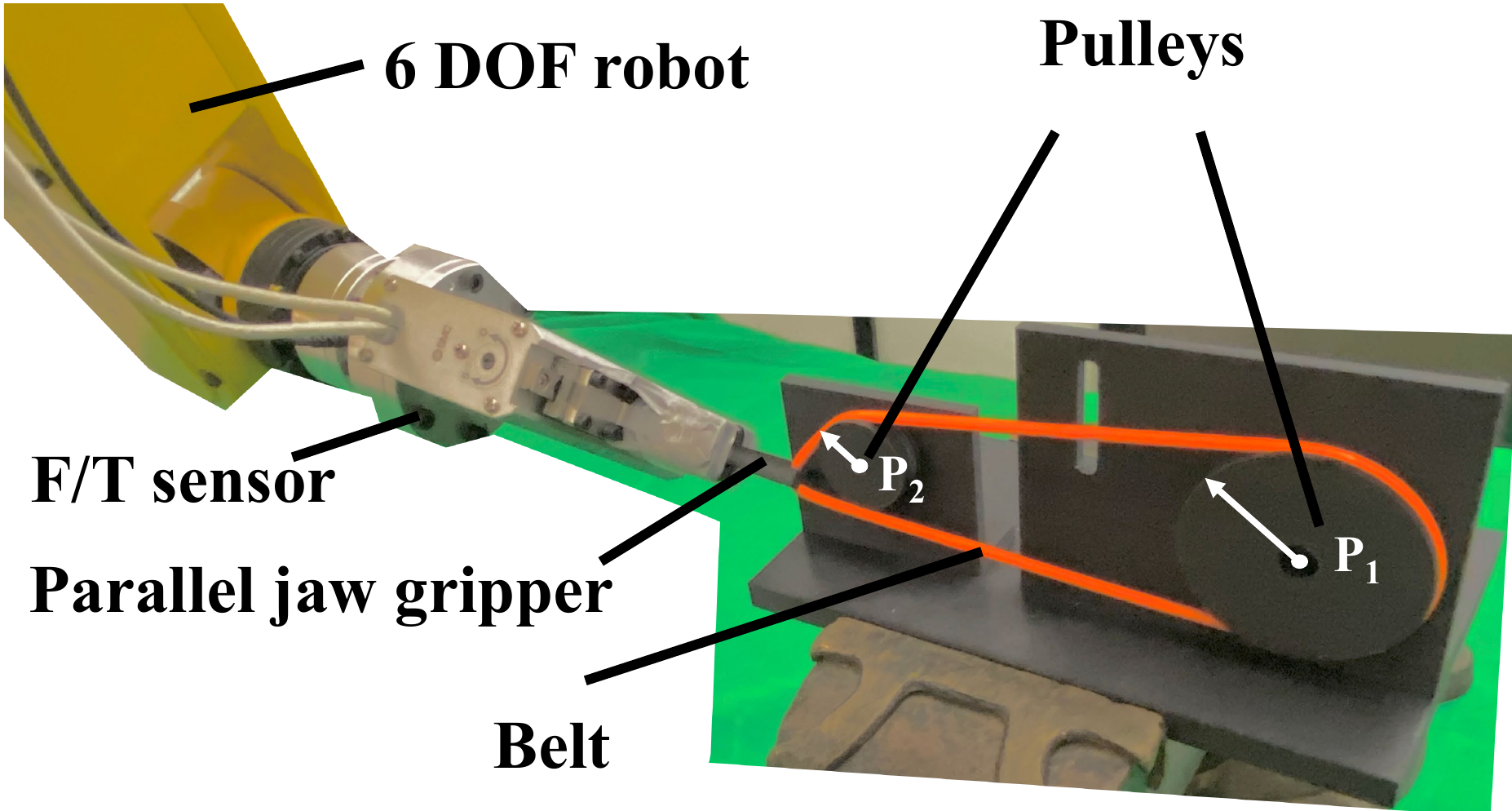}
    \caption{Real Setup of the Belt Drive Unit system.}
    \label{fig:BDU_setup}
\end{figure}
The real world system is represented in Figure~\ref{fig:BDU_setup} where the objective of the manipulation problem is to wrap the belt, held by a robotic manipulator around the two pulleys. The elastic belt is modeled through a 3D keypoint representation. The hybrid behavior of the model generated by the contacts between the belt and the pulleys and the elastic properties of the belt is captured by the complementarity constraints. 

\begin{figure}[h]
	\centering
    \includegraphics[scale=0.25]{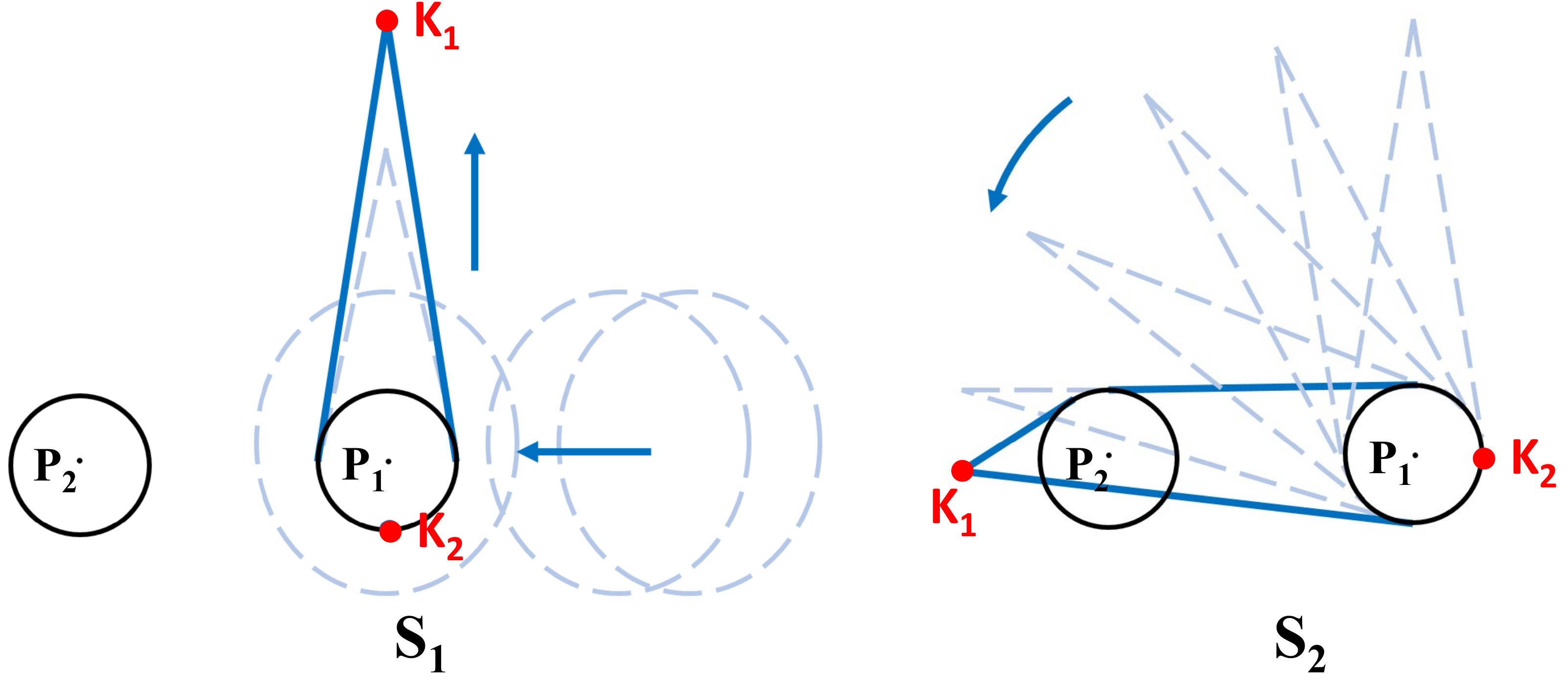}
    \caption{Visualization of the two subtasks decomposition, $S_1$ and $S_2$. $P_1$ and $P_2$ are two pulleys. The blue lines represent the belt gripped  by a robot at keypoint $K_1$, and $K_2$ is the lower keypoint. $S_1$: The belt wraps around the first pulley $P_1$ and it is stretched. $S_2$: The belt rotates around the first pulley and it is assembled onto the second pulley $P_2$.}
    \label{fig:BDU_subtasks}
\end{figure}

The full manipulation task has been divided into two subtasks as shown in Figure~\ref{fig:BDU_subtasks}. The goals of the first and second subtask are to wrap the belt around the first and second pulley, respectively.
We formulate two trajectory optimization problems one for each of the two subtasks as a MPCC of the form in~\eqref{dynopt}. 
\begin{figure}[!h]
    \centering
    \includegraphics[width=\columnwidth]{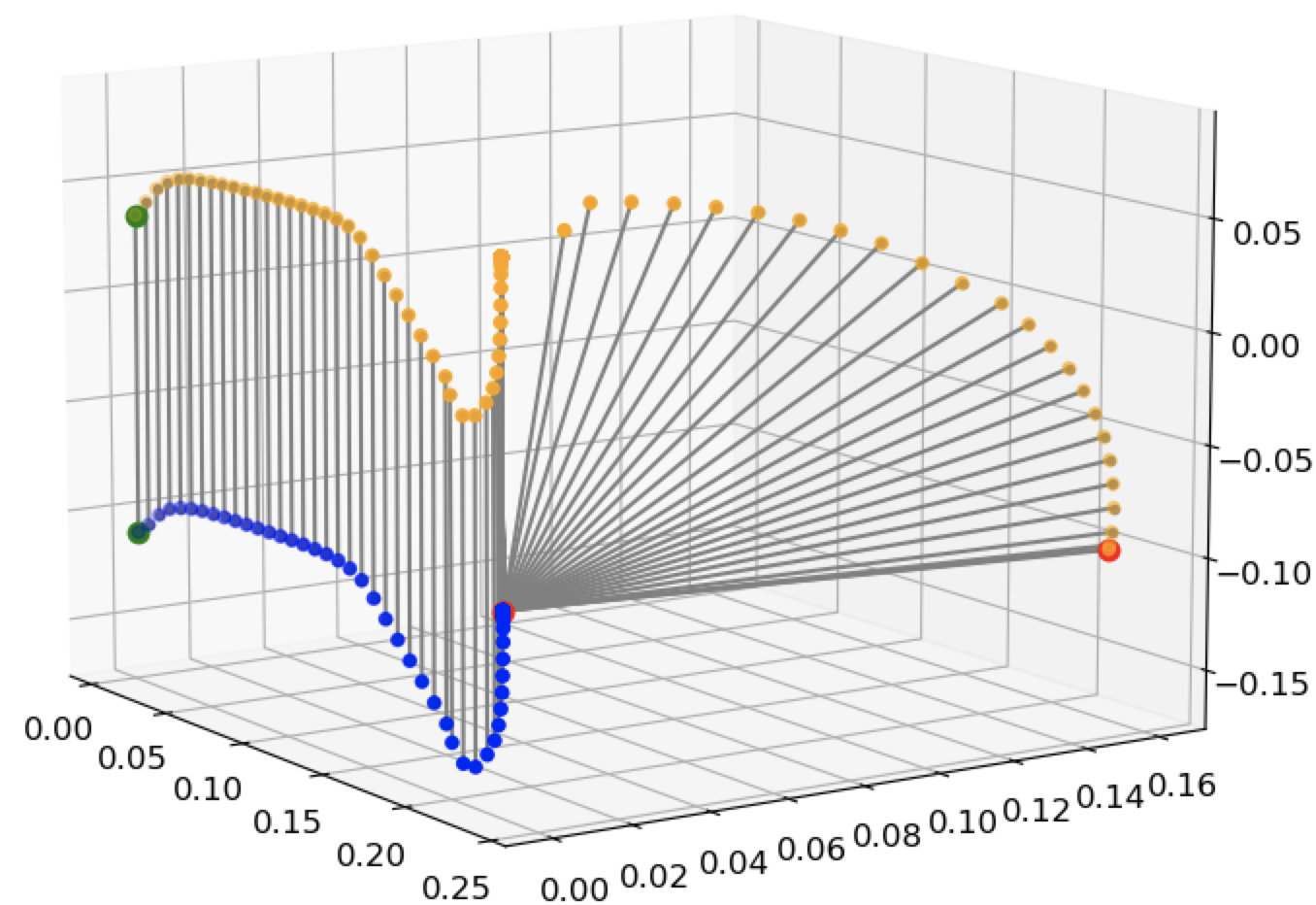}
    \caption{The optimal trajectory to assemble the belt is shown. The orange points represent the trajectory of the upper keypoint, $K_1$, and the blue points represent the lower keypoint, $K_2$, which together represent the model of the belt. The green and red points are the starting and the final points, respectively. The grey lines are virtual connections between $K_1$ and $K_2$ for illustration only. The belt approaches the first pulley (not shown) then there is a movement downwards to hook the pulley with the lower keypoint from below during subtask 1. The lower keypoint is then hooked onto the pulley and will not move. Then, the higher keypoint, $K_2$, moves toward the second pulley (not shown) stretching the belt and wraps around the pulley during subtask 2. }
    \label{fig:BDU_traj3D}
\end{figure}
%
% MPCC for BDU
% \begin{subequations}
% \begin{align}
%     \min\limits_{x,u, \lambda} &\, \sum_{k=0}^{N} (x(k)-x^{goal})^\top Q(x(k)-x^{goal}) +\\
%   &u(k)^TRu(k) + w(\bar{\lambda}_0(k) - \bar{\lambda}_0^{desired})^2 \label{opt:obj} \\
%     \text{s.t.} \quad &\, \dot{x} = Ax + Bu + G + f(x,\lambda) \label{opt:dynamic} \\
%                 &\, 0 \leq  \lambda_2 \quad \perp \quad  \lambda_0 \geq 0 \label{opt:ineq} \\
%                 &\, \lb{x} \leq x \leq \ub{x}, \, \lb{u} \leq u \leq \ub{u}, \, \lb{\lambda} \leq \lambda \leq \ub{\lambda} \label{opt:bnds}
% \end{align}\label{opt}
% \end{subequations}
% where... 
\begin{figure*}
	\begin{minipage}[]{1.0\columnwidth}
		\centering
		\includegraphics[scale=0.5]{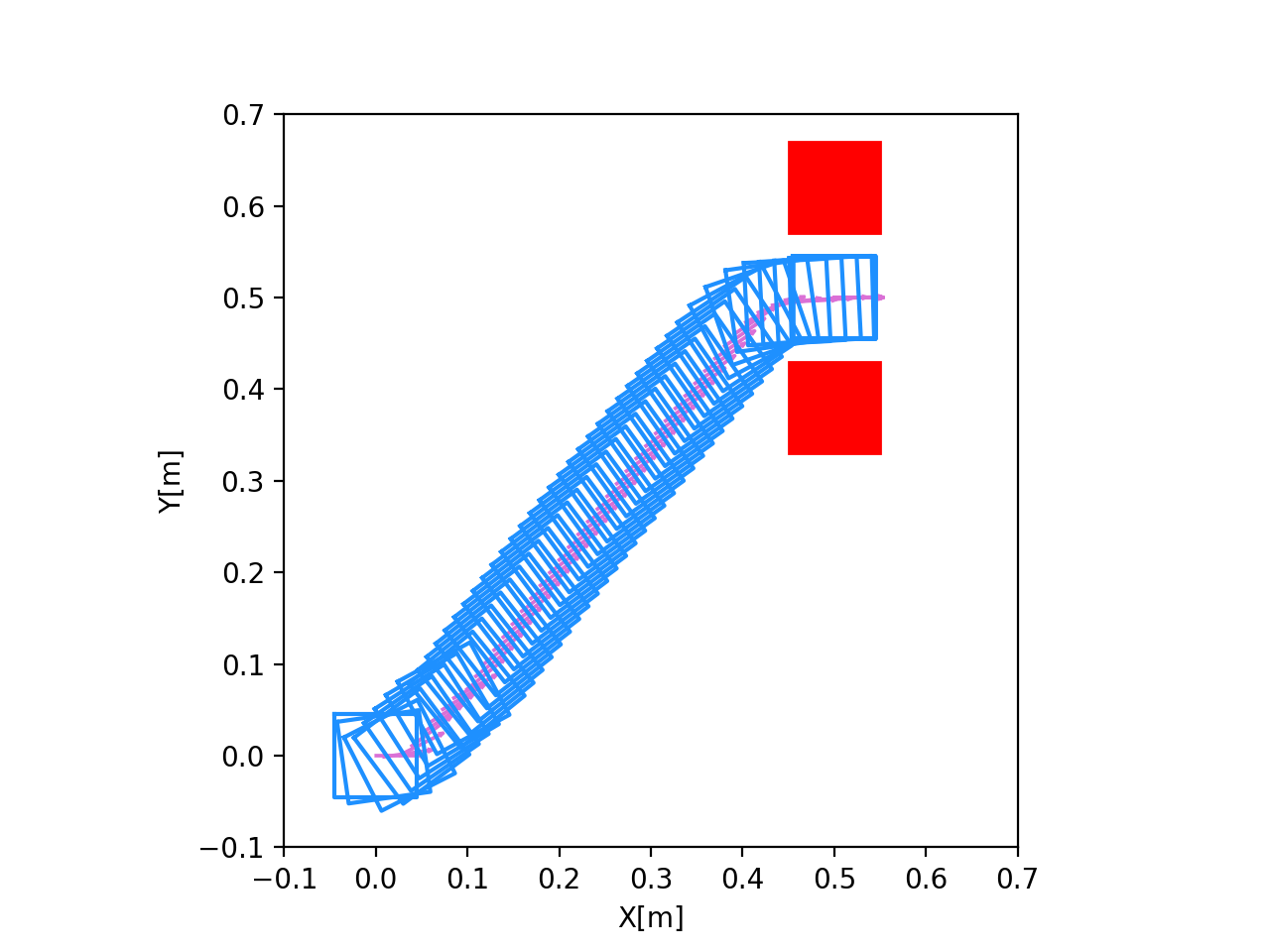}
		\vskip -0.05in
		\subcaption{Pushing sequence for initial position $(0,0,0)$ and desired goal $(0.5,0.5,0)$}\label{fig:pushing_obs_solution_ex1}
	\end{minipage}
	\begin{minipage}[]{1.0\columnwidth}
		\centering
		\includegraphics[scale=0.5]{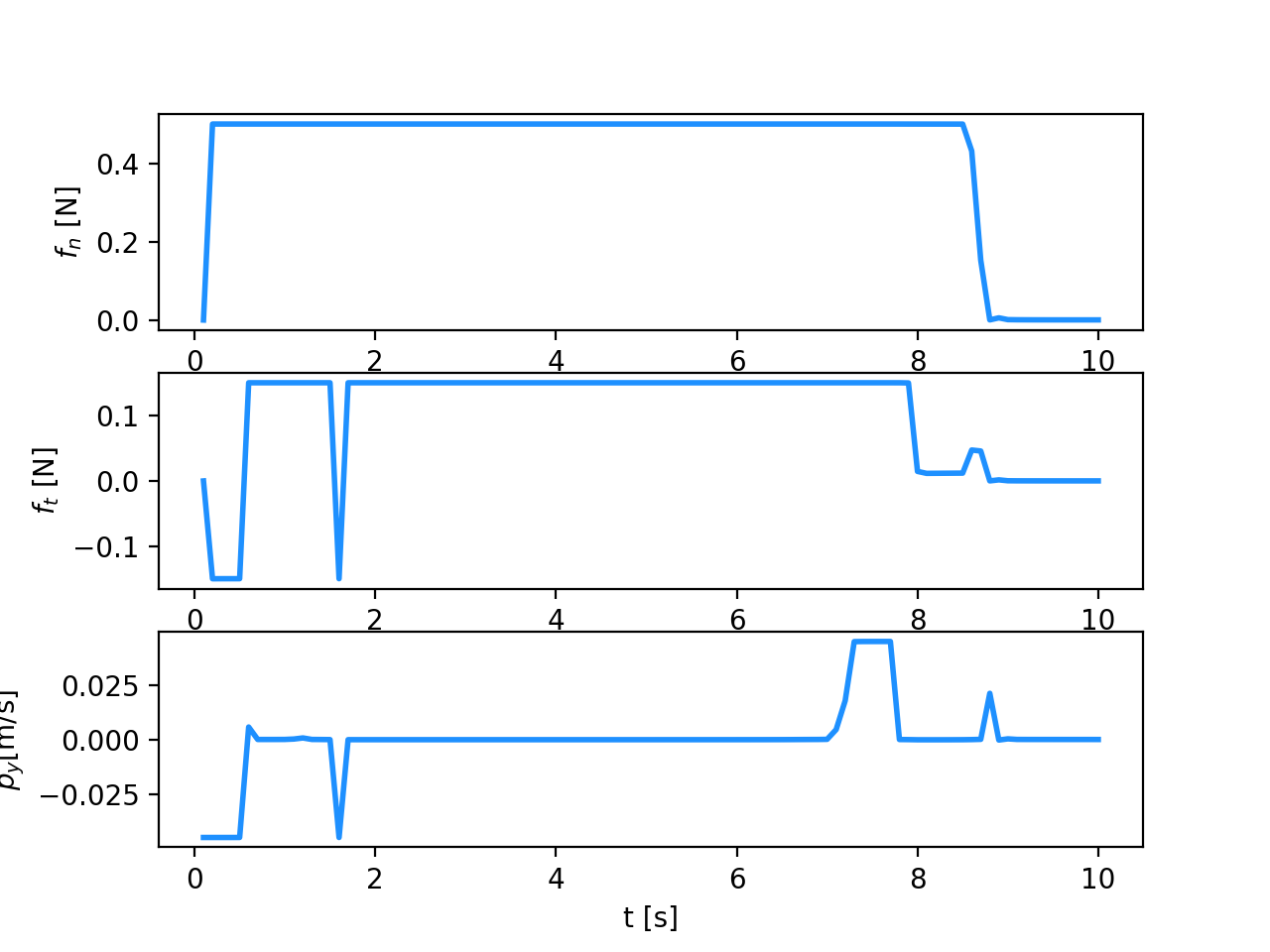}
		\vskip -0.05in
		\subcaption{Optimal Controls obtained for  Example~\ref{fig:pushing_obs_solution_ex1}}\label{fig:pushing_obs_solution_ex1_input}
	\end{minipage}
	\begin{minipage}[]{1.0\columnwidth}
		\centering
		\includegraphics[scale=0.5]{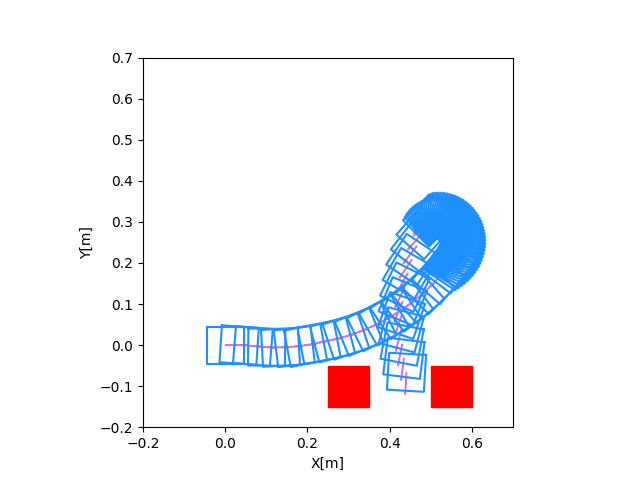}
		\vskip -0.05in
		\subcaption{Pushing sequence for initial position $(0,0,0)$ and desired goal $(0.45,-0.1,3\pi/2.)$}\label{fig:pushing_obs_solution_2}
	\end{minipage}
	\begin{minipage}[]{1.0\columnwidth}
		\centering
		\includegraphics[scale=0.5]{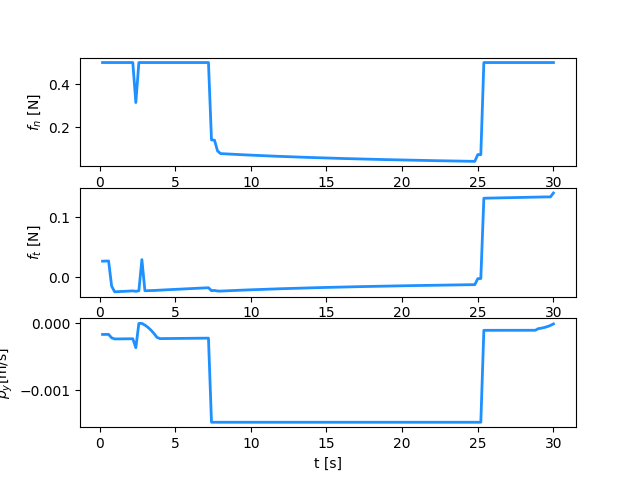}
		\vskip -0.05in
		\subcaption{Optimal Controls obtained for Example~\ref{fig:pushing_obs_solution_2}.}\label{fig:pushing_obs_solution_2_input}
	\end{minipage}
	\vskip -0.05in
    \caption{Planar pushing in the presence of obstacles. Our proposed formulation in \pyrobocop\ allows us to solve the collision avoidance. The trajectory of $\dot{p}_y$ shows the slipping contact sequence between the pusher and the slider. Pusher maintains a sticking contact when $\dot{p}_y=0$.}
	\label{fig:example_pushing_with_obstacles}
\end{figure*}

Details on the modeling assumptions, the division into the two subtasks, the exact formulation including the explanation of the dynamics and complementarity constraints can be found in our previous paper~\citep{jin2021trajectory}. In Figure~\ref{fig:BDU_traj3D} we report successful trajectories computed by \pyrobocop\  to assemble the belt drive unit combining the optimal trajectories obtained in the two subtasks. The optimal trajectory was implemented on the real system with a tracking controller, see~\citep{jin2021trajectory} for further details.

\subsection{Planar Pushing With Obstacles}

In this section, we show the solution to some planar pushing scenarios in the presence of obstacles and show that our proposed method can handle complementarity constraints as well as obstacle avoidance constraints simultaneously. We demonstrate our approach on two different pushing scenarios with same initial condition for the slider but different location of the obstacles and different goal states for the slider. In particular, the initial state of the slider in both these examples was set to $\mathbf{x}_{\texttt{init}}=(0,0,0)$ and the goal state for the two conditions was specified as $\mathbf{x}_g = (0.5,0.5,0)$ and $(-0.1,-0.1,3\pi/2)$. We add the obstacles next to the goal state so that \pyrobocop\ has to find completely different solution compared to the case when there are no obstacles. The initial point of contact between the pusher and the slider is $p_y=0$.
The optimal pushing sequence to reach the goal states for the slider are shown in  Figures~\ref{fig:pushing_obs_solution_ex1} and~\ref{fig:pushing_obs_solution_2}. To provide more insight about the solution, we also provide the plot of the input sequences in Figures~\ref{fig:pushing_obs_solution_ex1_input} and~\ref{fig:pushing_obs_solution_2_input}. The slipping contact sequence between the slider and the pusher is seen in the plot of $\dot{p}_y$. Sticking contact occurs when $\dot{p}_y=0$. We show that the proposed solver can optimize for the desired sequence of contact modes in order to reach the target state. For the example in Figure~\ref{fig:pushing_obs_solution_ex1}, the objective is a function of target state and control inputs. For the example in Figure~\ref{fig:pushing_obs_solution_2}, the Mayer objective function is used.

\subsection{Optimization with Mode Enumeration}

\begin{figure}
        \centering
        \begin{subfigure}[b]{0.495\textwidth}
            \centering
            \includegraphics[width=\textwidth]{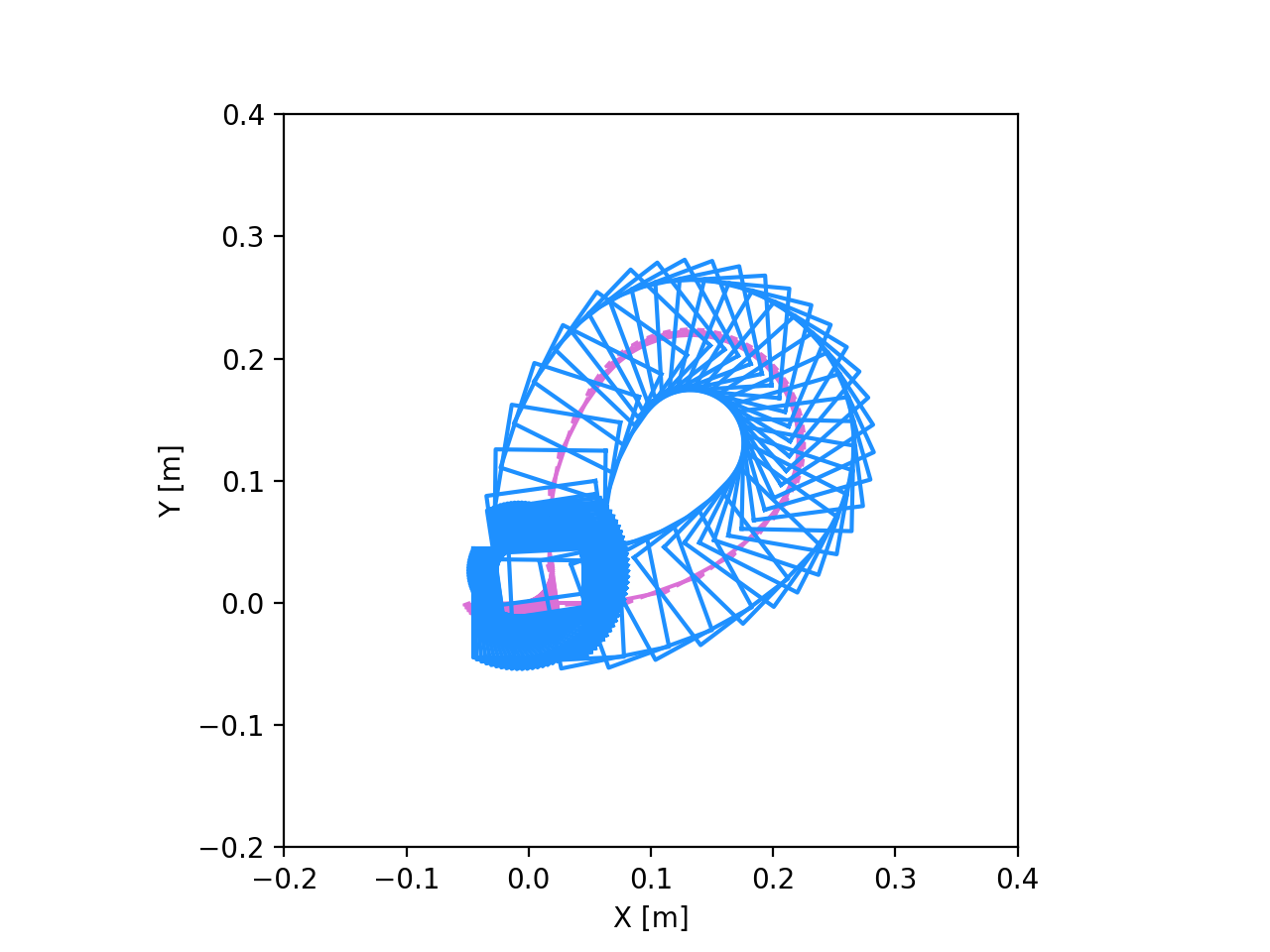}
            \caption{Optimal pushing sequence computed by \pyrobocop.}
            \label{fig:pushing_mode_seq}
        \end{subfigure}
        \hfill
        \begin{subfigure}[b]{0.495\textwidth}  
            \centering 
            \includegraphics[width=\textwidth]{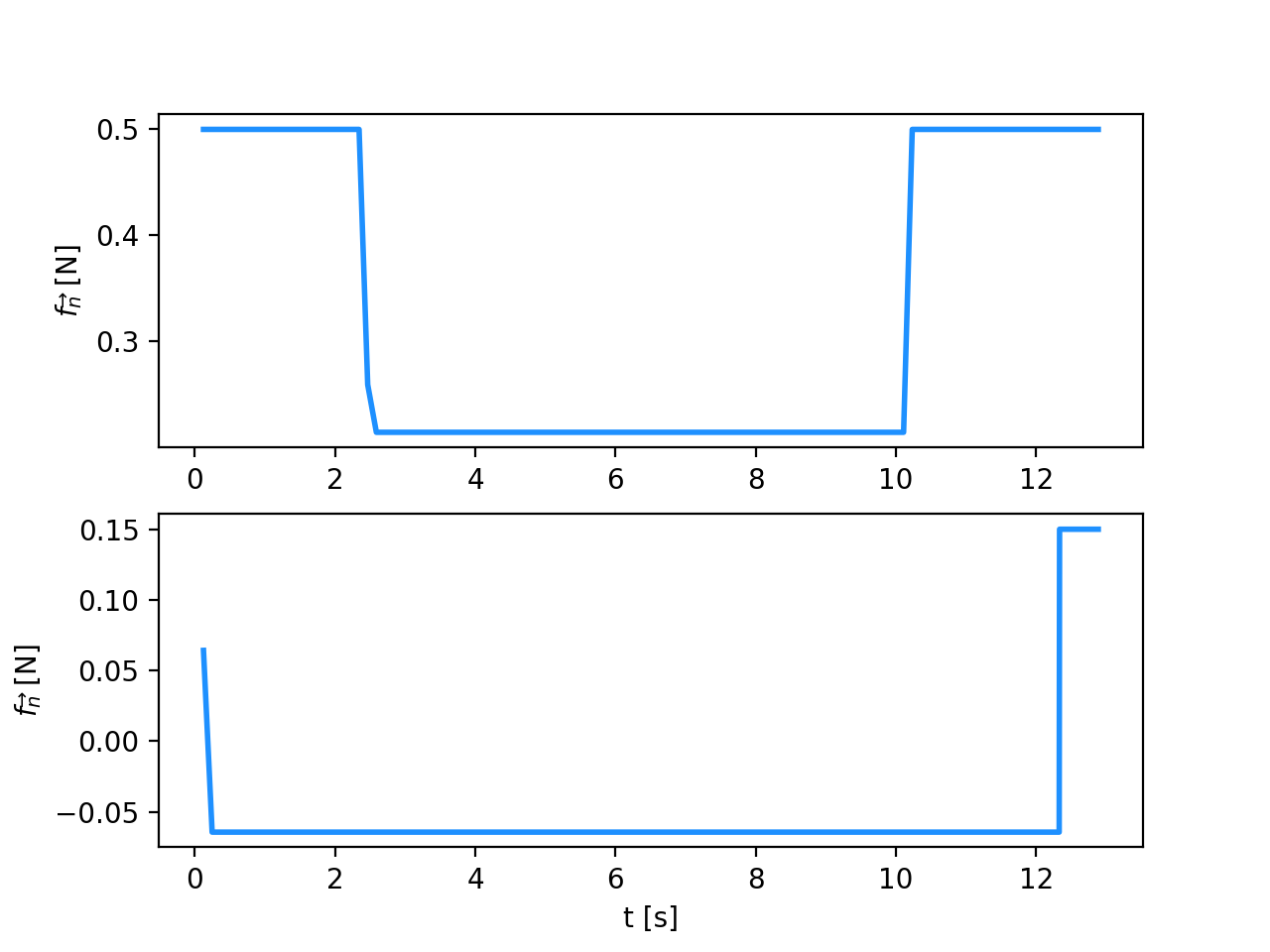}
            \caption{Optimal control inputs computed by \pyrobocop.}
            \label{fig:pushing_mode_seq_inputs}
        \end{subfigure}
        \caption{Optimal pushing sequence and control inputs obtained by optimizing mode sequence.}
        \label{fig:mode_seq_pushing}
\end{figure}

We show our approach of optimization over fixed mode sequences using the quasi-static pushing model which was presented in Section~\ref{subsec:manipulation_problem} while considering sticking contact at the $4$ faces of the slider (see Figures~\ref{fig:pushing} and~\ref{fig:pushing_analytical}). In particular, we use the dynamics model and the problem described in~\citep{doshi2020hybrid} to show solutions obtained by \pyrobocop\ in the case where the mode sequence is pre-specified. Note that this can be easily extended to the case where one can search for the mode-sequence using the approach discussed in~\citep{doshi2020hybrid}. Thus, we do not discuss mode sequence search here.  

The contact model in this case can be obtained from the model described in Section~\ref{subsec:manipulation_problem}, Eq~\ref{eqn:pushing_dynamics} with $\dot{p}_y=0$. Thus we only consider sticking contact between the pusher and the slider. The modes appear based on which face the pusher contacts with the slider, and thus we have four different modes that could be used during any interaction. For a given mode, state-space of the pusher-slider system is then $3$ dimensional while the input is only $2$ dimensional. We use our formulation presented in Section~\ref{subsec:mode_sequences} to solve for the optimization problem with pre-specified mode sequence. The optimization process ensures continuity of dynamics and selection of final time for each mode in a trajectory. The initial state of the slider is $\mathbf{x}_{\texttt{init}}= (0,0,0)$ and the goal state of the slider is $\mathbf{x_g}=(0,0,\pi)$. The two modes we use for this example are pushing from the left face followed by pushing from the top face of the slider. The trajectory obtained by \pyrobocop\ is shown in Figure~\ref{fig:pushing_mode_seq}. The inputs used in different modes is shown in Figure~\ref{fig:pushing_mode_seq_inputs}. The objective function used is the Mayer objective function, i.e., the minimum-time problem. The time spent in mode $1$ is $12.36$ seconds and in mode $2$ is $0.56$ seconds.

% \begin{figure}
%     \centering
%     \includegraphics[scale=0.60]{figures/pushing_mode_seq.png}
%     \caption{Planar pushing with a given mode sequence where the mode represents the point of contact between the pusher and the slider.}
%     \label{fig:pushing_mode_seq}
% \end{figure}

\subsection{Trajectory Optimization with a Machine Learning Model}
\begin{figure}
    \centering
    \includegraphics[scale=0.50]{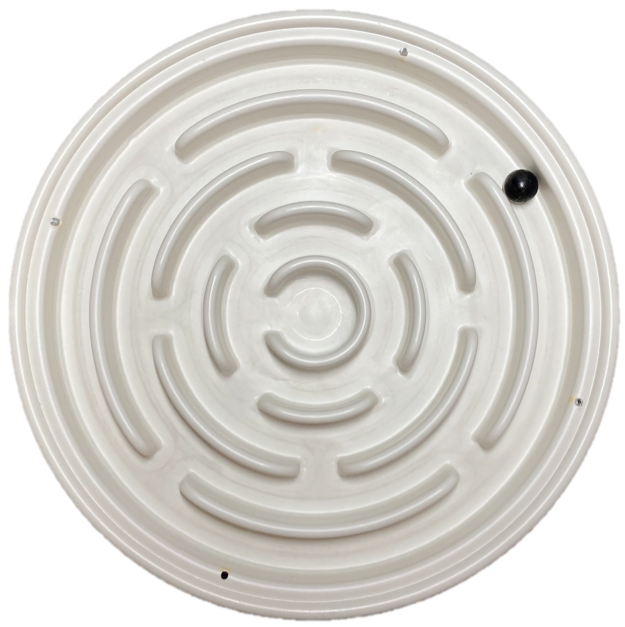}
    \caption{The circular maze environment for which we show trajectories using the learned Gaussian process models.}
    \label{fig:real_maze}
\end{figure}

%\subsection{Maze}
We illustrate the usage of \pyrobocop\ to control a complex dynamical system such as the circular maze represented in Figure~\ref{fig:real_maze}. The  goal in this system is to tip and tilt the maze in order to move a marble from an outer ring into the inner-most ring. The movement of the maze is actuated by two servomotors. The forward dynamics of the marble moving in the maze are learned using Gaussian Process Regression, as described in our previous work \citep{romeres2019semiparametrical}. The model is assumed to be a black box system and \adolc\ is used to compute the derivatives. \pyrobocop\ computes the control sequence and the marble's trajectory to reach one of the gates at each of the $4$ rings.
\begin{figure}[h]
\centering
\includegraphics[width=0.475\textwidth]{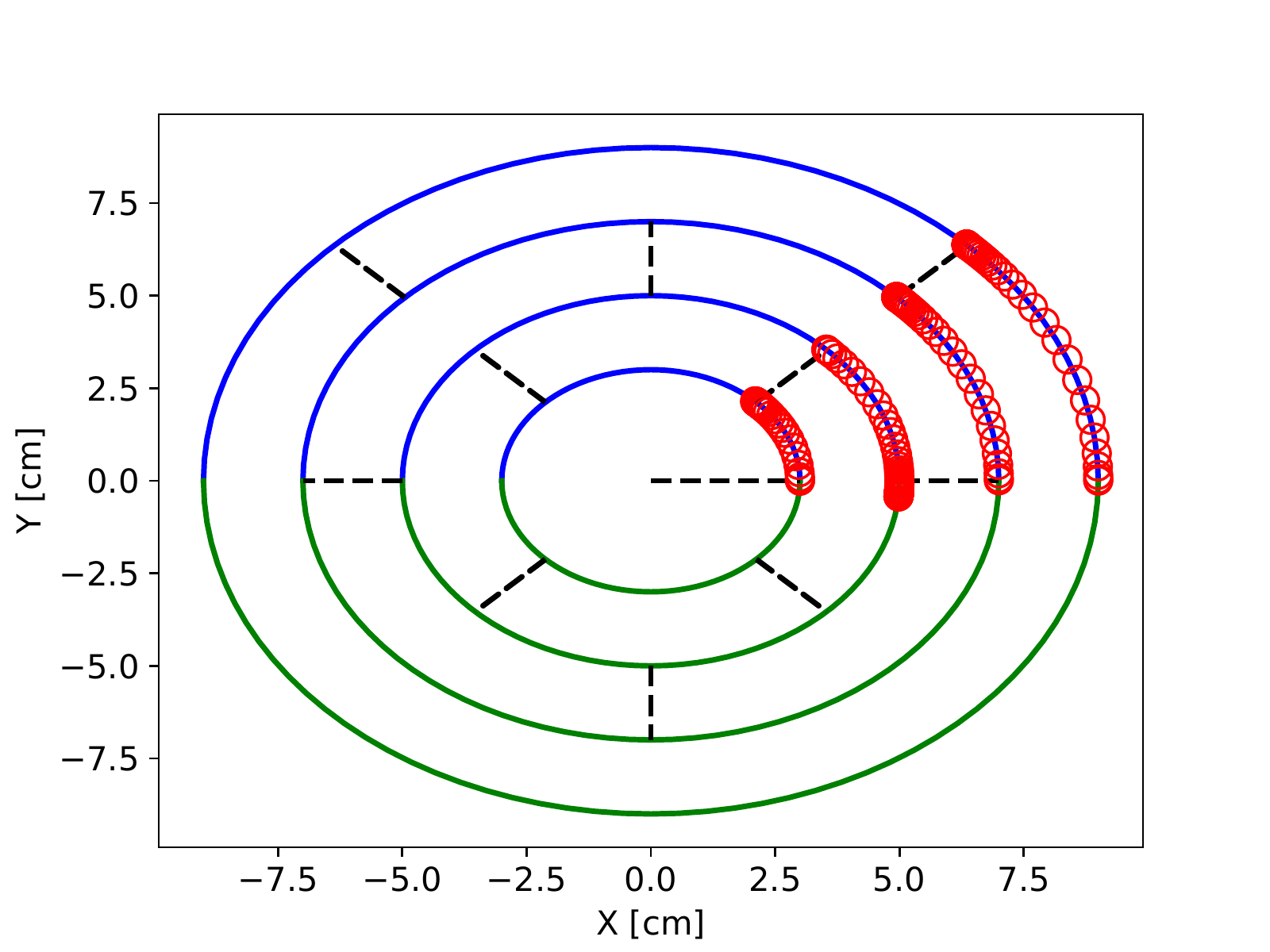}
\caption{The red circles represent the marble's optimal trajectory computed by \pyrobocop\ for each of the 4 rings of the maze.}
\label{fig:maze_traj}
\end{figure}
Figure~\ref{fig:maze_traj} shows the computed optimal trajectory of the marble in each of the rings, illustrated on a schematic of the maze.

% \begin{tabular}{|l|l|l|l|l|}
%   \hline
%   Time [s] & Ring 1 & Ring 2 & Ring 3 & Ring 4 \\
%   \hline
%   Ipopt & 55.71 & 0.48 &  0.535 &  0.371 \\
%   \hline
%   Func Eval & 53.29 & 147.56 &  142.04 &  64.89 \\
%   \hline
%   Objective & 53.29 & 0.49 &  0.37 &  0.35 \\
%   \hline
% \end{tabular}

\begin{figure}[h]
    \centering
    \includegraphics[width=0.5\textwidth]{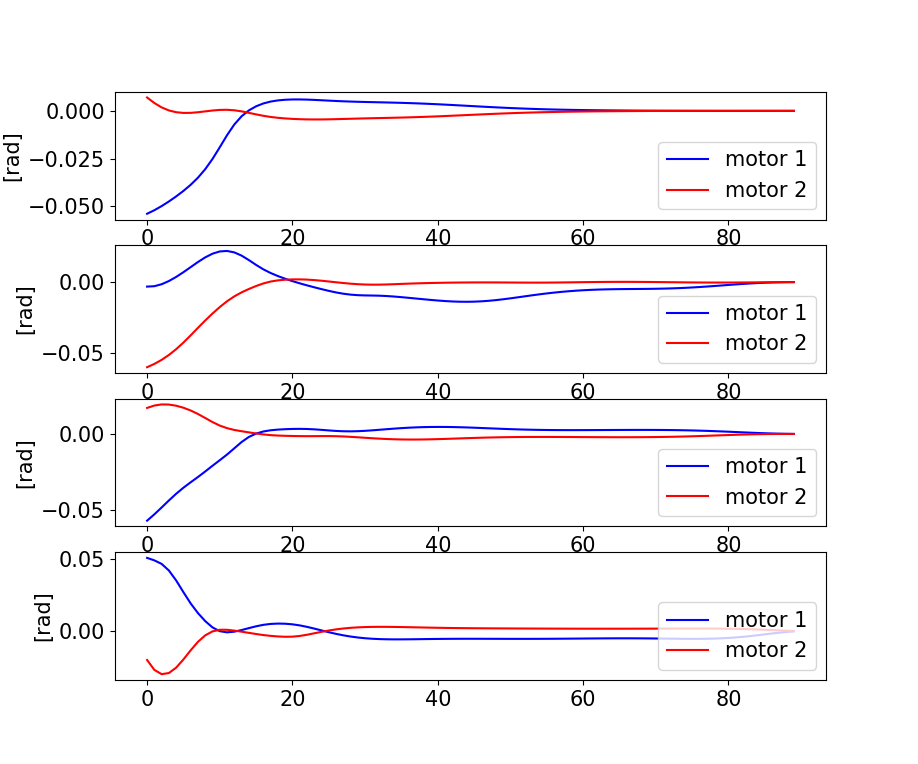}
    \caption{Optimal control sequence for the maze system in each of the 4 rings.}
    \label{fig:control_maze}
\end{figure}

The optimal control sequence in each of the ring is shown in Figure~\ref{fig:control_maze}.

%\diego{add as a remark?}
\begin{remark}
The choice of using \adolc\ gives the opportunity of having any machine learning model written in standard python and the computation of the derivatives comes automatic together with the non-zero sparsity structure without requiring any insight of the model itself. However, this does not preclude the option that if the derivatives are available from other toolboxes e.g., pytorch these derivatives can also be provided to the solver instead than the ones coming from \adolc\ .
\end{remark}

%---------------------------------COMPARISON-------------%

\subsection{Comparison with \casadi\  and \pyomo}
The purpose of this section is to compare \pyrobocop\ with some state-of-the-art open-source software for optimization and control, namely \casadi\ and \pyomo\ . These two software define their own syntax to model complex nonlinear optimization problems. In particular, \casadi\ uses a specific symbolic language that allows a fast automatic differentiation and translation to C-code, while \pyomo\ offers automatic differentation via the AMPL Solver Library. In \pyrobocop\ we provide to the user both an interface to specify the model and constraints derivatives manually and an automatic differentiation capability by using \adolc\ which computes the derivatives in C-code and provides access to the sparsity pattern which can be used during optimization. Furthermore, neither \casadi\ nor \pyomo\ offer specific constructors to handle complementarity constraints which is one of the major focuses in \pyrobocop. 

We compared the three packages on three different benchmark dynamical systems with increasing state dimension: an inverted pendulum, an acrobot and a quadrotor. The purpose is to see if \pyrobocop\ converges to the same solutions in comparable time. For fair comparison the models, the constraints and the initial conditions are identical in all three software and we used IPOPT as the solver in all the experiments. We solve the OCP for these systems $5$ times using each software and record the mean and standard deviation for solution time.  Results are shown in Table~\ref{tab:comparison_table}. 
\begin{figure}[h]
        \centering
            \includegraphics[width=0.475\textwidth]{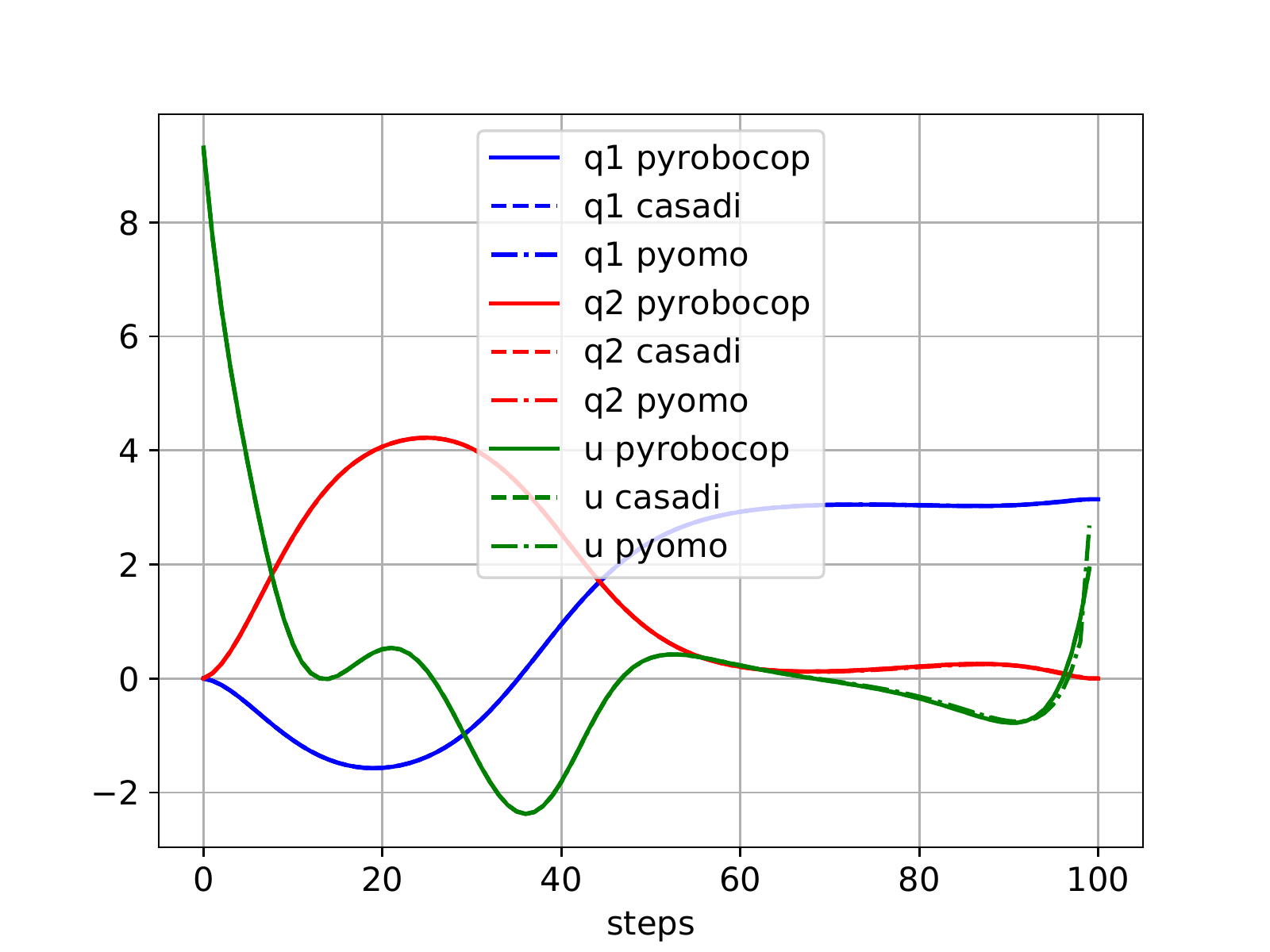}
            \caption{Optimal trajectory computed by the three optimization packages.}
            \label{fig:traj_acrobot}
    \end{figure}
    
% \begin{figure}
%         \centering
%         \begin{subfigure}[b]{0.475\textwidth}
%             \centering
%             \includegraphics[width=\textwidth]{figures/traj_pendulum.pdf}
%             %\caption{Optimal trajectory computed by the three optimization packages.}
%             \label{fig:ur10_boxplot1}
%         \end{subfigure}
%         \hfill
%         \begin{subfigure}[b]{0.475\textwidth}  
%             \centering 
%             \includegraphics[width=\textwidth]{figures/control_pendulum.pdf}
%             %\caption{Comparison in data efficiency among the 3 best models with confidence intervals.}
%             \label{fig:ur10_boxplot2}
%         \end{subfigure}
%         \caption{Optimal trajectory and control input computed by the three optimization packages.}
%         \label{fig:ur10_fd_performance}
%     \end{figure}

% add number of iterations
\begin{table*}
    \centering
    \begin{tabular}{|c|c|c|c|c|c|c|}
  \hline
  System & State & IPOPT & Func Eval & Problem & Cost & Iterations\\
 & Dim & Time & Time & size & Func &\\
  \hline
  Pendulum \pyrobocop\ & 2 & 0.112$\pm$ 0.003 & 0.101 $\pm$ 0.003 & 1048 & 19.89 & 29 \\
  \hline
  Pendulum \casadi\ & 2 & 0.058 $\pm$ 0.001 & 0.016 $\pm$ 0.0008  & 750 & 19.89 & 21 \\
  \hline
  Pendulum \pyomo\ & 2 & 0.146 $\pm$ 0.008 & 0.008 $\pm$ 0.0008  & 755 & 20.13 & 31\\
  \hline
  Acrobot \pyrobocop\ & 4 & 2.282 $\pm$ 0.05 & 1.85 $\pm$ 0.019 & 1296 & 62.724 & 349 \\
  \hline
  Acrobot \casadi\ & 4 & 1.175 $\pm$0.023  & 0.706 $\pm$ 0.008  & 900 & 62.52 & 355 \\
  \hline
  Acrobot Pyomo & 4 & 2.374 $\pm$ 0.039 & 0.652 $\pm$ 0.021 & 909 & 62.76 & 265\\
  \hline
  Quadrotor \pyrobocop\ & 12 & 0.871 $\pm$ 0.016 & 0.786 $\pm$ 0.013  & 7988  & 156.01 & 21\\
  \hline
  Quadrotor \casadi\ & 12 & 0.353 $\pm$ 0.002 &  0.050 $\pm$ 0.001 & 5600 & 156.01 & 9\\
  \hline
  Quadrotor \pyomo\ & 12 & 1.190 $\pm$ 0.038 & 0.057 $\pm$ 0.001  & 5628 & 155.64 & 26\\
  \hline
\end{tabular}

    \caption{Comparison between \pyrobocop , \casadi\ and \pyomo\ on three non-linear systems of different dimensions. In all cases \pyrobocop\ achieves comparable performance with both \casadi\ and \pyomo.}
    \label{tab:comparison_table}
\end{table*}
% \begin{tabular}{|c|c|c|c|c|c|}
%   \hline
%   System & State & IPOPT & Func Eval & Problem & Cost \\
%  & Dim & Time & Time & size & Func \\
%   \hline
%   Pendulum (ours) & 2 & 0.177 & 0.092 & 1048 & 19.89\\
%   \hline
%   Pendulum Casadi & 2 & 0.072 & 0.014  & 1050 & 19.89 \\
%   \hline
%   Pendulum Pyomo & 2 & 0.254 & 0.008  & 755 & 20.13 \\
%   \hline
%   Acrobot (ours) & 4 & 2.540 & 1.881 & 1296 & 62.724\\
%   \hline
%   Acrobot Casadi & 4 & 2.340  & 0.565  & 900 & 62.52\\
%   \hline
%   Acrobot Pyomo & 4 & 4.792 & 0.799 & 909 & 62.76 \\
%   \hline
%   Quadrotor (ours) & 12 & 0.807 & 0.729  & 7988  & 156.01 \\
%   \hline
%   Quadrotor Casadi & 12 & 0.342 &  0.049 & 5600 & 156.01 \\
%   \hline
%   Quadrotor Pyomo & 12 & 1.174 & 0.056  & 5628 & 155.64 \\
%   \hline
% \end{tabular}

In the table we can first notice that in each experiment the three software converge to a solution with a similar cost function implying convergence to similar optimal solution. An example of this is shown for the acrobot system in Figure~\ref{fig:traj_acrobot}, where both the optimal state trajectory and the optimal control sequence computed by the three software are plotted, and there is an almost exact overlap. Similar plots have been obtained for the other systems but are not shown for sake of brevity. Second, we can notice that the problem size, which is the total number of variables in the optimization problem is higher in \pyrobocop\ w.r.t. the other software. We have chosen a formulation of the transcription that depends only on the number of collocation points $n_c$ and independent of the choice of the roots. With the choice of \texttt{legendre} roots in the collocation we observe that the number of variables in \pyomo\, is larger than that in \pyrobocop\, which remains invariant to choice of roots. The analysis of the difference is orthogonal to the main point of this section. It can also be observed that the number of iterations are different in the three optimization packages. This can be attributed to the difference in the problem size for the underlying NLP, which can also be seen in Table~\ref{tab:comparison_table}.
%In our formulation of the NLP problem~\eqref{nlp} the $i-$th finite element $i=\{1,\ldots,N_e\}$ is associated with the variables $[x_{i,0}, \{x_{i,j},\dot{x}_{i,j},y_{i,j},u_{i,j}\}_{j=1}^{n_c}]\in \R^{n_d+(2n_d + n_a + n_u)n_c}$. The additional variables are introduced to ensure that continuity equations can be written  \diego{but some ore not variables because of the final bounds?} while in \adolc\ and \pyomo\ ..... this gives as the advantage that ... However, the drawback is in the decrease of speed performance..\diego{not really true w.r.t. pyomo}

We can conclude that all the three software packages are more or less equivalent in solving the considered systems. Based on our experiments, it seems that \casadi\ tends to outperform the other two in terms of computational time. The main advantages of \pyrobocop\ over the other packages are in offering convenient in-built methods to (i) model  complementarity constraints and (ii) automatically formulate the collision avoidance constraints for all pairs of user-specified objects using the novel obstacle avoidance formulation.

\section{Concluding Remarks}\label{sec:conclusions}
This paper presented \pyrobocop\ which is a python-based optimization package for model-based control of robotic systems. This package has been developed with the motivation to allow python-based control of contact-rich systems operating in constrained environments in the presence of other obstacles. We showed that \pyrobocop\ can be used to solve trajectory optimization problems of a number of dynamical systems in different configurations such as with contact and collision avoidance constraints. We demonstrated two practical scenarios of planar pushing and belt-drive unit assembly where one needs to consider the collision avoidance as well as contact constraints. A description of the functions that a potential user needs to implement in order to solve their control or optimization problem has been provided. The software has been benchmarked against two other SOTA optimization packages to verify the solution quality and timing obtained by \pyrobocop.
\subsection{Strengths of \pyrobocop}
\pyrobocop\ is a model-based trajectory and control package for systems with non-linear and non-smooth dynamics. In particular, \pyrobocop\ can handle systems with contact as well as collision constraints with a novel complementarity formulation. \pyrobocop\ also allows automatic differentiation by using \adolc. To the best of our knowledge, \pyrobocop\ is the only python-based, open-source software that allows handling of contact \& collision constraints and automatic differentiation for control and optimization. Unlike most of the competing optimization solvers which are available in python, \pyrobocop\ allows users to provide dynamics information in python through a simple script using Numpy data structures~\citep{harris2020array}. We show that we achieve similar computation times as achieved by other SOTA optimization toolboxes like \casadi\ and \pyomo. Note that these solvers do not provide adaptive relaxations for  complementarity constraints as in \pyrobocop. Another advantage is that \pyrobocop\ allows interfacing to MPCC solver using standard Numpy data structures instead of using data structures designed for \pyrobocop. This makes \pyrobocop\ easier to use when compared to other packages like \casadi\ and \pyomo. \pyrobocop\ also simplifies solution of collision avoidance problems by requiring users to only provide the bounding polytope for each obstacle. A potential user does not need to specify the constraints arising from collision avoidance -- this is handled by \pyrobocop\ internally. This also reduces the risk of modeling errors made by users when defining these constraints whicah can be very hard to track down.
\subsection{Limitations of \pyrobocop\ and Future Work} We would also like to highlight some of the limitations of \pyrobocop. Since \pyrobocop\ uses \ipopt\ as the solver for the resulting MPCC problems, it borrows limitations of \ipopt. In particular, one of the main limitations is that \pyrobocop\ can find only local solutions. Furthermore, it might require good initialization to find even the local solutions. Furthermore, it can not detect infeasibility of the underlying optimization problem provided by the user. Another possible limitation is given by interfacing \pyrobocop\ with \adolc. While, as described above, this is one of the strengths of \pyrobocop\, it also carries some limitations as we still have to rely on an external code to do the automatic differentiation while other software like \casadi\ have built-in source code transformation into C and can handle the differentiation internally with faster performance. In the future we will explore other open source options. 

While we have shown that \pyrobocop\ can handle a broad range of robotic control problems, the current software has been developed under the premise that the dynamics is provided to \pyrobocop\ as DAEs, which requires expert knowledge for system specification. We identify this is a limitation as this restricts the developmental usage of \pyrobocop. In the future, we would work towards integrating \pyrobocop\ with a physics engine for easy specification of dynamics to provide solution to complex manipulation tasks with real-time feedback~\citep{dong2021icra}. We will also extend the MPCC formulation presented in the current paper to allow finite horizon model predictive control (MPC) for hybrid systems in future research to allow real-time control of contact-rich systems.

%\appendix{}
%\input{appendix}

\bibliographystyle{SageH}
\bibliography{reference}

\end{document}